%% file: main.tex
\UseRawInputEncoding
\documentclass[10pt,twocolumn,letterpaper]{article}

\usepackage[pagenumbers]{cvpr} 

\usepackage{graphicx}
\usepackage{amsmath}
\usepackage{amssymb}
\usepackage{booktabs}

\usepackage[dvipsnames, svgnames, x11names]{xcolor}
\usepackage[english]{babel}
\usepackage{tablefootnote}
\usepackage{enumitem}
\usepackage{amsthm}
\usepackage{tabu}
\usepackage{times}
\usepackage{epsfig}
\usepackage{multirow}
\usepackage{tabularx}
\usepackage{eqparbox} 
\usepackage{colortbl}

\usepackage{makecell}
\usepackage{pifont} 
\usepackage{dsfont}  

\usepackage{balance}

\definecolor{citecolor}{HTML}{0071bc}
\usepackage[
    pagebackref,
    breaklinks=true,
    letterpaper=true,
    colorlinks,
    bookmarks=true
]{hyperref}

\newcommand{\pgraph}[1]{\noindent\textbf{#1.\,}}
\input{section/math_commands.tex}

\usepackage[capitalize]{cleveref}
\crefname{section}{Sec.}{Secs.}
\Crefname{section}{Section}{Sections}
\Crefname{table}{Table}{Tables}
\crefname{table}{Tab.}{Tabs.}

\def\cvprPaperID{6318} 
\def\confName{CVPR}
\def\confYear{2023}

\begin{document}

\title{MixPHM: Redundancy-Aware Parameter-Efficient Tuning for\\Low-Resource Visual Question Answering}

\author{Jingjing Jiang \qquad 
Nanning Zheng\thanks{Corresponding author.}
\\ 
Institute of Artificial Intelligence and Robotics, Xi'an Jiaotong University\\ 
{\tt\small jingjingjiang2017@gmail.com\qquad nnzheng@mail.xjtu.edu.cn} 
}

\maketitle

\begin{abstract} 
Recently, finetuning pretrained Vision-Language Models (VLMs) has been a prevailing paradigm for achieving state-of-the-art performance in Visual Question Answering (VQA). However, as VLMs scale, finetuning full model parameters for a given task in low-resource settings becomes computationally expensive, storage inefficient, and prone to overfitting. Current parameter-efficient tuning methods dramatically reduce the number of tunable parameters, but there still exists a significant performance gap with full finetuning. In this paper, we propose \textbf{MixPHM}, a redundancy-aware parameter-efficient tuning method that outperforms full finetuning in low-resource VQA. Specifically, MixPHM is a lightweight module implemented by multiple PHM-experts in a mixture-of-experts manner. To reduce parameter redundancy, MixPHM reparameterizes expert weights in a low-rank subspace and shares part of the weights inside and across experts. Moreover, based on a quantitative redundancy analysis for adapters, we propose \textbf{Redundancy Regularization} to reduce task-irrelevant redundancy while promoting task-relevant correlation in MixPHM representations. Experiments conducted on VQA v2, GQA, and OK-VQA demonstrate that MixPHM outperforms state-of-the-art parameter-efficient methods and is the only one consistently surpassing full finetuning. 
\end{abstract}

\input{section/intro.tex}

\input{section/rw.tex}

\input{section/method.tex}

\input{section/exp.tex}

\section{Conclusion and Limitation} 
In this paper, we propose a redundancy-aware parameter-efficient tuning method to adapt pretrained VLMs to the low-resource VQA task. 
The proposed MixPHM reduces task-irrelevant redundancy while promoting task-relevant correlation through a proposed redundancy regularization. 
Experiments demonstrate its effectiveness and superiority in terms of performance and parameter efficiency.

Redundancy is a double-edged sword. 
In addition to reducing task-irrelevant redundancy, we can also exploit task-relevant redundancy already learned by pretrained VLMs to enhance performance. 
Although MixPHM emphasizes reducing task-irrelevant redundancy, there is no explicit guarantee that the reduced redundancy is ineffective for given tasks. 
As such, a potential prospect is to investigate how to explicitly delimit and minimize task-irrelevant redundancy.

{\small 
\pgraph{Acknowledgements}
This work was supported by the National Science Foundation of China (Grant No. 62088102). 
}

\input{section/supp.tex}

{\small
\bibliographystyle{ieee_fullname}
\bibliography{ref}
}


\end{document}

%% file: section/math_commands.tex
\newcommand{\Transpose}{\mathrm{T}}
\newcommand{\std}[1]{_{\pm{#1}}}

\newcommand{\bv}{\boldsymbol{b}}

\newcommand{\hv}{\boldsymbol{h}}

\newcommand{\xv}{\boldsymbol{x}}
\newcommand{\yv}{\boldsymbol{y}}

\newcommand{\thetav}{\boldsymbol{\theta}}

\newcommand{\Amat}{\mathbf{A}}

\newcommand{\Smat}{\mathbf{S}}
\newcommand{\Tmat}{\mathbf{T}}
\newcommand{\Umat}{\mathbf{U}}

\newcommand{\Wmat}{\mathbf{W}}

\newcommand{\Zmat}{\mathbf{Z}}


%% file: section/intro.tex
\section{Introduction}

\begin{figure}[!t]
\centering
\includegraphics[width=0.8\linewidth]{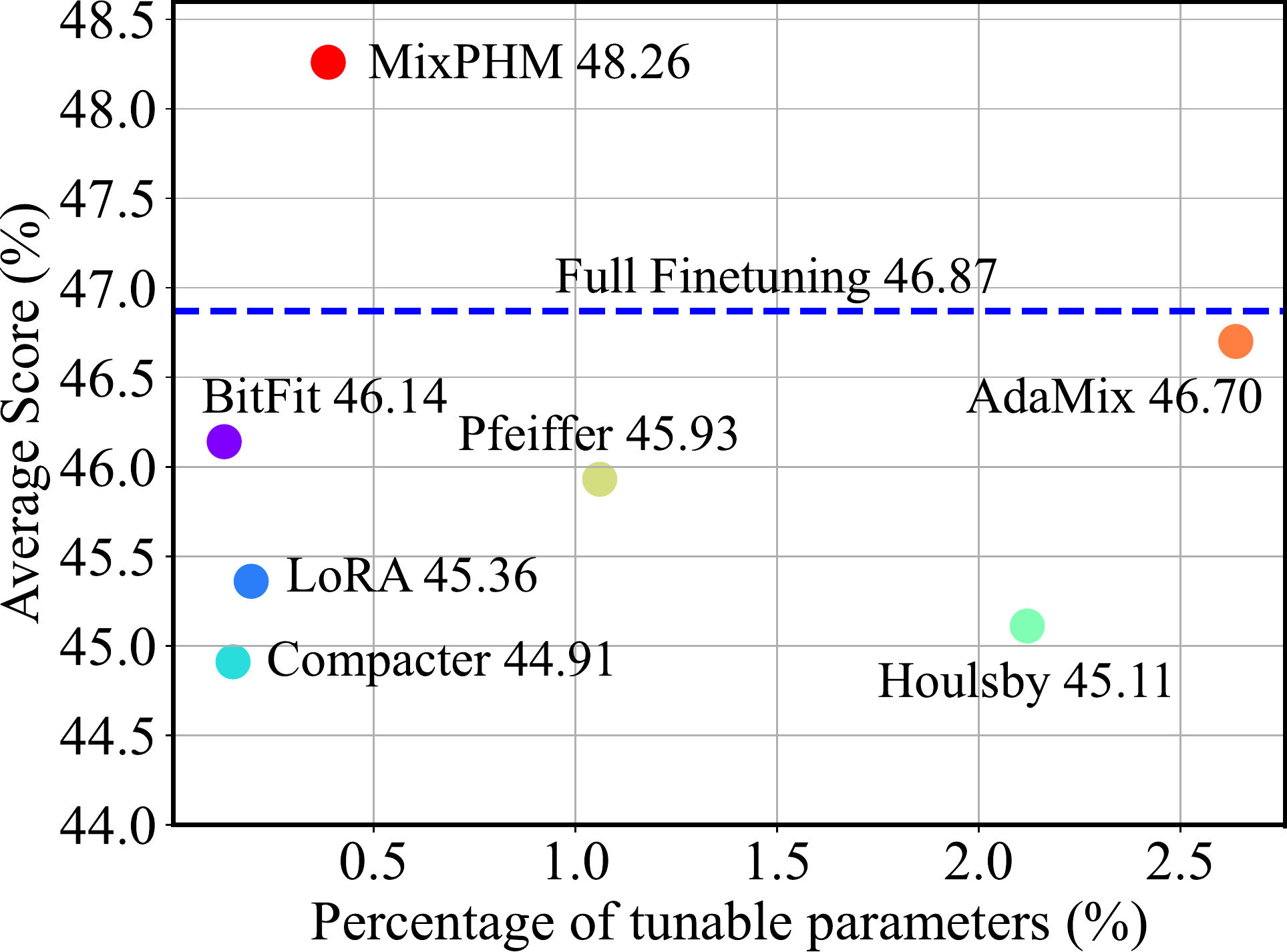}
\vspace{-3mm}
\caption{
\textbf{Comparison between parameter-efficient methods.}
In a low-resource setting (\ie, with 64 training samples), we show the average score across five seeds on VQA v2 (y-axis) and the percentage of tunable parameters \wrt pretrained VL-T5 (x-axis). 
}
\label{fig:adapter_acc}
\vspace{-4mm}
\end{figure}

Adapting pretrained vision-language models (VLMs)~\cite{cho2021unifying,li2021align,zeng2022multi,kim2021vilt,dou2022empirical,li2022blip,wang2022ofa} to the downstream VQA task~\cite{antol2015vqa} through finetuning has emerged as a dominant paradigm for achieving state-of-the-art performance. 
However, as the scale of VLMs continues to expand, finetuning an entire pretrained model that consists of millions or billions of parameters engenders a substantial increase in computational and storage costs, and exposes the risk of overfitting (poor performance) in low-resource learning. 
Parameter-efficient tuning methods, only updating newly-added lightweight modules (\eg, Houlsby~\cite{houlsby2019parameter}, Pfeiffer~\cite{pfeiffer2020adapterfusion}, Compacter~\cite{karimi2021compacter}, and AdaMix~\cite{wang2022adamix}) or updating a tiny number of parameters in pretrained models (\eg, BitFit~\cite{zaken2022bitfit} and LoRA~\cite{hu2022lora}), are thus proposed to handle these challenges.

However, as illustrated in \Cref{fig:adapter_acc}, the aforementioned parameter-efficient tuning methods substantially reduce the number of tunable parameters, but their performance still lags behind full finetuning. 
Among these methods, adapter-based approaches~\cite{houlsby2019parameter,pfeiffer2020adapterfusion,karimi2021compacter,wang2022adamix} allow for more flexible parameter sharing~\cite{sung2022vl} and are more storage-efficient by storing only a copy of these adapters instead of the entire pretrained model. 
In particular, AdaMix~\cite{wang2022adamix} employs adapters as experts within a mixture-of-experts (MoE)~\cite{shazeer2017outrageously} architecture to enhance adapter capacity and achieves comparable performance to full finetuning. 
Nevertheless, this MoE mode leads to an increase in the number of tunable parameters due to \textit{parameter redundancy} between adapters.

In this paper, we build upon adapter-based approaches to explore more parameter-efficient tuning methods that can outperform full finetuning on low-resource VQA. Specifically, when adapting pretrained VLMs to a given task, we focus on two key improvements: 
\textbf{(\emph{i})} \textit{Reducing parameter redundancy while  maintaining adapter capacity}. 
Striking a balance between parameter efficiency and model capacity is crucial since an excessive reduction of tunable parameters may lead to underfitting, limiting the ability of adapters to capture sufficient task-relevant information~\cite{karimi2021compacter}. 
\textbf{(\emph{ii})} \textit{Reducing task-irrelevant redundancy while promoting task-relevant correlation in representations}. 
In practice, adapters leverage residual connections to integrate task-specific information learned from a target dataset and prior knowledge already implied in pretrained VLMs. 
However, recent works~\cite{jiang2022finetuning,mahabadi2021variational,wang2021infobert} have highlighted that pretrained models inevitably contain redundant and irrelevant information for target tasks, resulting in statistically spurious correlations between representations and labels, thereby hindering performance and generalization~\cite{tishby2015deep,wang2022rethinking}. 
To enhance the effectiveness of adapters, our objective thus is to maximize the acquisition of task-relevant information while discarding task-irrelevant information from pretrained VLMs.

To this end, we propose \textbf{MixPHM}, a redundancy-aware parameter-efficient tuning method that can efficiently decrease tunable parameters and reduce task-irrelevant redundancy while promoting task-relevant correlation in representations. 
MixPHM is implemented with multiple PHM-experts in a MoE manner. 
To reduce \textbf{(\emph{i})} \textit{parameter redundancy}, MixPHM first decomposes and reparameterizes expert weights into a low-rank subspace. 
Then, it further reduces tunable parameters and facilitates information transfer via global and local weight sharing. 
To achieve the improvement \textbf{(\emph{ii})}, we first quantify the redundancy for adapters in representation spaces. 
The result reveals that adapter representations are redundant with pretrained VLM representations, but exhibit limited correlation to the final task-used representations. 
Motivated by this finding, we propose \textbf{Redundancy Regularization}, which is incorporated into MixPHM and serves to reduce \textit{task-irrelevant redundancy} by decorrelating the similarity matrix between MixPHM representations and pretrained VLM representations. 
Simultaneously, it promotes \textit{task-relevant correlation} by maximizing the mutual information between MixPHM representations and the final task-used representations.

We conduct experiments on VQA v2~\cite{goyal2017making}, GQA~\cite{hudson2019gqa}, and OK-VQA~\cite{marino2019ok}, and demonstrate that the proposed MixPHM consistently outperforms full finetuning and state-of-the-art parameter-efficient tuning methods when adapting pretrained VLMs to low-resource VQA. 
In summary, our contributions are as follows: 
(1) We propose MixPHM, a redundancy-aware parameter-efficient tuning method for adapting pretrained VLMs to downstream tasks. 
(2) We propose Redundancy Regularization, a novel regularizer to reduce task-irrelevant redundancy while promoting task-relevant correlation in MixPHM representations. 
(3) Extensive experiments show MixPHM achieves a better trade-off between performance and parameter efficiency, leading to a significant performance gain over existing methods.

%% file: section/rw.tex
\section{Related Work}


\pgraph{Vision-Langauge Pretraining} 
Vision-language pretraining~\cite{tan2019lxmert,gan2020large,huang2021seeing,zhang2021vinvl,jia2021scaling,kim2021vilt,li2021align,dou2022empirical,zhong2022regionclip} aims to learn task-agnostic multimodal representations for improving the performance of downstream tasks in a finetuning fashion. 
Recently, a line of research~\cite{cho2021unifying,li2022blip,li2021align,hu2021unit,wang2022ofa} has been devoted to leveraging encoder-decoder frameworks and generative modeling objectives to unify architectures and objectives between pretraining and finetuning. 
VLMs with an encoder-decoder architecture generalize better. 
In this paper, we explore how to better adapt them to low-resource VQA~\cite{antol2015vqa}.

\pgraph{Parameter-Efficient Tuning} 
Finetuning large-scale pretrained VLMs on given downstream datasets has become a mainstream paradigm for vision-language tasks. However, finetuning the full model consisting of millions of parameters is time-consuming and resource-intensive. 
Parameter-efficient tuning~\cite{zaken2022bitfit,mahabadi2021parameter,ruckle2021adapterdrop,mao2022unipelt,he2022towards,yang2022robust,zhang2022differentiable} vicariously tunes lightweight trainable parameters while keeping (most) pretrained parameters frozen, which has shown great success in NLP tasks. 
According to whether new trainable parameters are introduced, these methods can be roughly categorized into two groups: (1) tuning partial parameters of pretrained models, such as BitFit~\cite{zaken2022bitfit} and FISH Mask~\cite{sung2021training}, (2) tuning additional parameters, such as prompt (prefix)-tuning~\cite{lester2021power,li2021prefix}, adapter~\cite{houlsby2019parameter,pfeiffer2020adapterfusion}, and low-rank methods~\cite{karimi2021compacter,hu2022lora}. 

Motivated by the success in NLP, some works~\cite{lin2022adapt,zhang2022hyperpelt,sung2022vl} have begun to introduce parameter-efficient methods to tune pretrained VLMs for vision-language tasks. 
Specifically, 
Lin~\etal~\cite{lin2022adapt} investigate action-level prompts for vision-language navigation. 
VL-Adapter~\cite{sung2022vl} extends adapters to transfer VLMs for various vision-language tasks. 
HyperPELT~\cite{zhang2022hyperpelt} is a unified parameter-efficient framework for vision-language tasks, incorporating adapter and prefix-tuning. 
In addition, Frozen~\cite{tsimpoukelli2021multimodal} and PICa~\cite{yang2022empirical} use prompt-tuning techniques~\cite{lester2021power} to transfer the few-shot learning ability of large-scale pretrained language models to handle few-shot vision-language tasks. 
FewVLM~\cite{jin2022good} designs hand-crafted prompts to finetune pretrained VLMs for low-resource adaptation. 
In contrast, low-rank methods are more parameter-efficient but are rarely explored.

\begin{figure*}[!t]
\centering
\includegraphics[width=0.95\linewidth]{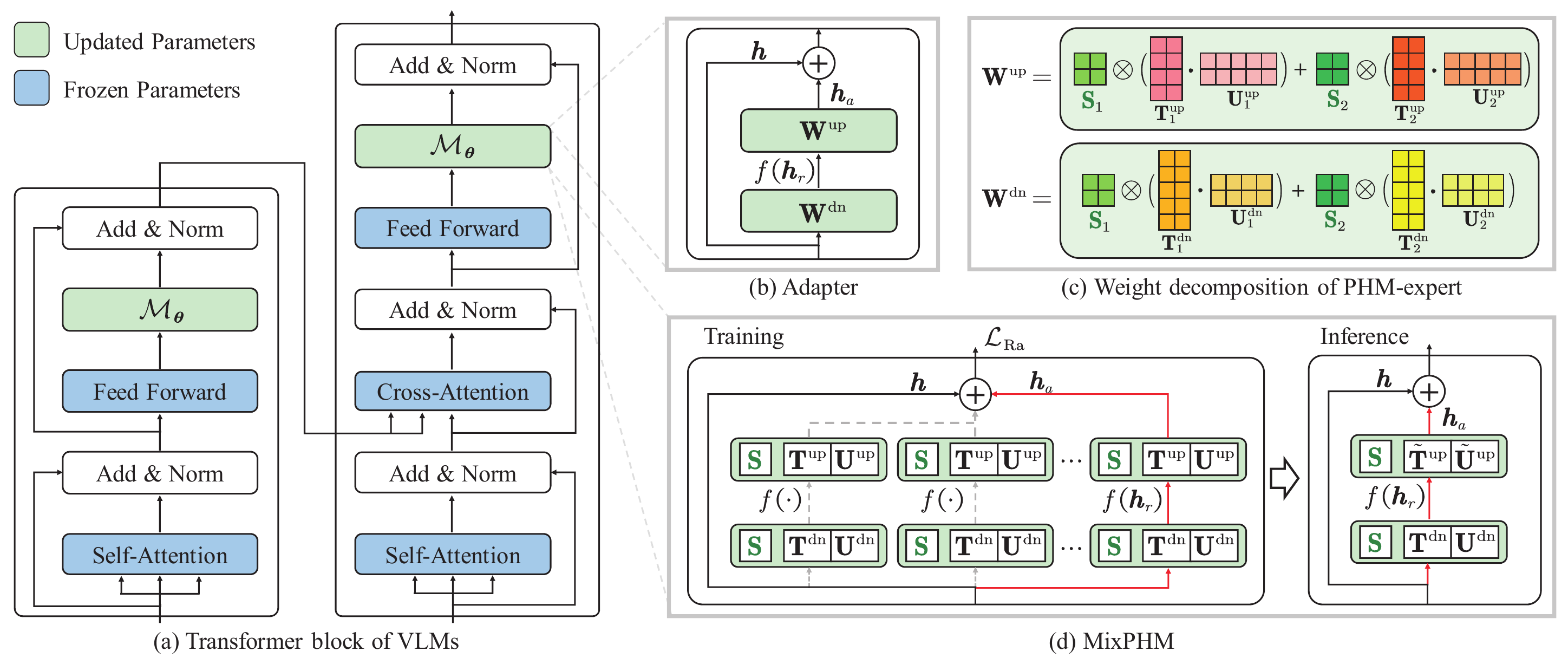}
\vspace{-3mm}
\caption{\textbf{Illustration of MixPHM} inserted into (a) one transformer block of VLMs. 
(b) The structure of standard adapter. 
(c) An example of the weight matrix decomposition in \cref{eq:ada_lphm} for a PHM-expert (here $n=2$, $d=10$, $d_r=8$, $d_k=2$). 
(d) MixPHM architecture with $N_e=3$ PHM-experts. 
During training, MixPHM randomly activates one PHM-expert to learn robust representations and exploits the proposed redundancy regularization $\mathcal{L}_{\text{Ra}}$ to reduce task-irrelevant redundancy while promoting task-relevant correlation. 
}
\label{fig:overall}
\vspace{-3mm}
\end{figure*}

\pgraph{Mixture-of-Experts} 
MoE~\cite{shazeer2017outrageously} aims to scale up model capacities and keep computational efficiency with conditional computation. 
Most recent works~\cite{lepikhin2021gshard,fedus2021switch,roller2021hash,lewis2021base,riquelme2021scaling,du2022glam} investigate how to construct large-scale vision or language transformer models using MoE and well-designed routing mechanisms in the pretraining stage. 
Despite its success in pretraining, MoE has not been widely explored in parameter-efficient tuning. 
MPOE~\cite{gao2022parameter} and AdaMix~\cite{wang2022adamix} are two recent works that tune pretrained language models by MoE. 
Specifically, MPOE considers additional FFN layers as experts and decomposes weight matrices of experts with MPO. 
AdaMix treats the added adapters as experts and increases adapter capacity by a stochastic routing strategy.

%% file: section/method.tex
\section{Preliminary}
\label{sec:preli}


\pgraph{Problem Definition} 
We follow recent work~\cite{cho2021unifying,wang2022simvlm} to formulate VQA as a generative modeling task, \ie, generating free-form textual answers for a given question instead of selecting a specific one from the predefined set of answers. 
Formally, we denote a VQA dataset with $\mathcal{D}=\{(I, Q, y)\in \mathcal{I}\times \mathcal{Q}\times \mathcal{Y} \}$, where $I$ is an image, $Q$ is a question, and $y$ is an answer. 
Assuming that a given pretrained VLMs $\mathcal{M}_{\Theta}$ is parameterized by a tiny number of tunable parameters $\thetav$, the general problem of adapting pretrained VLMs for VQA is to tune $\mathcal{M}_{\thetav}$ in a parameter-efficient manner on $\mathcal{D}$.

\pgraph{Mixture-of-Experts} 
A standard MoE~\cite{shazeer2017outrageously} is implemented with a set of $N_e$ experts $\{E_i\}_{i=1}^{N_e}$ and a gating network $G$. 
Each expert is a sub-neural network with unique weights and can learn from a task-specific subset of inputs. 
The gate conditionally activates $N_a$ ($1\le N_a \le N_e$) experts. 
Formally, given an input representation $\xv \in \mathbb{R}^{d}$, the $i$-th expert maps $\xv$ into $d_e$-dimensional space, \ie, $E_i(\cdot): \xv \rightarrow \mathbb{R}^{d_e}$, the gate generates a sparse ${N_e}$-dimensional vector, \ie, $G(\cdot): \xv \rightarrow \mathbb{R}^{N_e}$.
Then, the output $\yv \in \mathbb{R}^{d_e}$ of the MoE can be formulated as 
\begin{equation} 
\begin{aligned} 
\yv = \sum_{i=1}^{N_e} G(\xv)_i E_i(\xv),
\end{aligned}
\label{eq:moe} 
\end{equation}
where, $G(\xv)_i$ denotes the probability of assigning $\xv$ to the $i$-th expert, satisfying $\sum_{i=1}^{N_e} G(\xv)_i = 1$.

\pgraph{Parameterized Hypercomplex Multiplication} 
The PHM layer~\cite{zhang2021beyond} aims to generalize hypercomplex multiplications to fully-connected layer by learning multiplication rules from data. 
Formally, for a fully-connected layer that transforms an input $\xv \in \mathbb{R}^{d}$ to an output $\yv \in \mathbb{R}^{d_e}$, \ie, 
\begin{equation} 
\begin{aligned} 
\yv = \Wmat^{\Transpose}\xv + \bv, 
\end{aligned}
\label{eq:fc} 
\end{equation}
where, $\Wmat \in \mathbb{R}^{d \times d_e}$. 
In PHM, the weight matrix $\Wmat$ is learned via the summation of $n$ Kronecker products between $\Smat_j \in \mathbb{R}^{n \times n}$ and $\Amat_j \in \mathbb{R}^{\frac{d}{n} \times \frac{d_e}{n}}$:
\begin{equation} 
\begin{aligned} 
\Wmat = \sum_{j=1}^{n} \Smat_j \otimes \Amat_j, 
\end{aligned}
\label{eq:phm} 
\end{equation}
where, the hyperparameter $n \in \mathbb{Z}_{> 0}$ controls the number of the above summations, 
$d$ and $d_e$ are divisible by $n$, and $\otimes$ indicates the Kronecker product that generalizes the vector outer products to higher dimensions in real space. 
For example, the Kronecker product between $\Smat \in \mathbb{R}^{m \times k}$ and $\Amat \in \mathbb{R}^{p \times q}$ is a block matrix $\Smat \otimes \Amat \in \mathbb{R}^{mp \times kq}$, \ie, 
\begin{equation} 
\begin{aligned} 
\Smat \otimes \Amat = 
\begin{bmatrix}
s_{11}\Amat &\cdots &s_{1k}\Amat \\
\vdots &\ddots &\vdots \\
s_{m1}\Amat &\cdots &s_{mk}\Amat 
\end{bmatrix}, 
\end{aligned}
\label{eq:kp} 
\end{equation}
where, $s_{ij}$ denotes the element of matrix $\Smat$ at the $i$-th row and $j$-th column. 
As a result, replacing a fully-connected layer with PHM can reduce the trainable parameters by at most $1/n$ of the fully-connected layer.

\section{Methodology}
\label{sec:method}

We propose MixPHM, a redundancy-aware parameter-efficient tuning method to adapt pretrained VLMs. 
This section first analyzes the redundancy in the adapter representation space toward low-resource VQA (\cref{sec:revist}). 
Then, we sequentially detail architecture (\cref{sec:mixphm}), redundancy regularization (\cref{sec:L_ra}), and inference (\cref{sec:inference}) of MixPHM.

\subsection{Rethinking Redundancy in Adapter}  
\label{sec:revist}

As shown in \Cref{fig:overall} (b), adapter~\cite{houlsby2019parameter} is essentially a lightweight module, usually implemented by a two-layer feed-forward network with a bottleneck, a nonlinear function, and a residual connection. 
When finetuning pretrained VLMs on downstream tasks, adapters are inserted between the transformer layers of VLMs, and only the parameters of the newly-added adapters are updated, while the original parameters in pretrained VLMs remain frozen. 
Formally, given an input representation $\hv \in \mathbb{R}^d$, the down-projection layer $\Wmat^{\text{dn}} \in \mathbb{R}^{d \times d_r}$ maps $\hv$ to a lower-dimensional space specified by the bottleneck dimension $d_r$, \ie, $\hv_r \in \mathbb{R}^{d_r}$. 
The up-projection layer $\Wmat^{\text{up}} \in \mathbb{R}^{d_r \times d}$ maps $\hv_r$ back to the input size, \ie, $\hv_a \in \mathbb{R}^d$. 
Considering the residual and nonlinear function $f$, an adapter is defined as
\begin{align} 
&\hv_a = f(\hv \Wmat^{\text{dn}}) \Wmat^{\text{up}}, \\
&\hv \leftarrow \hv_a + \hv.  
\label{eq:adapter} 
\end{align}
Ideally, by incorporating task-specific information learned from a given downstream dataset ($\hv_a$) and prior knowledge already encoded in pretrained VLMs ($\hv$), adapters can quickly transfer pretrained VLMs to new tasks without over-parameterization or under-parameterization.

\begin{figure}[t]
\centering
\includegraphics[width=1.0\linewidth]{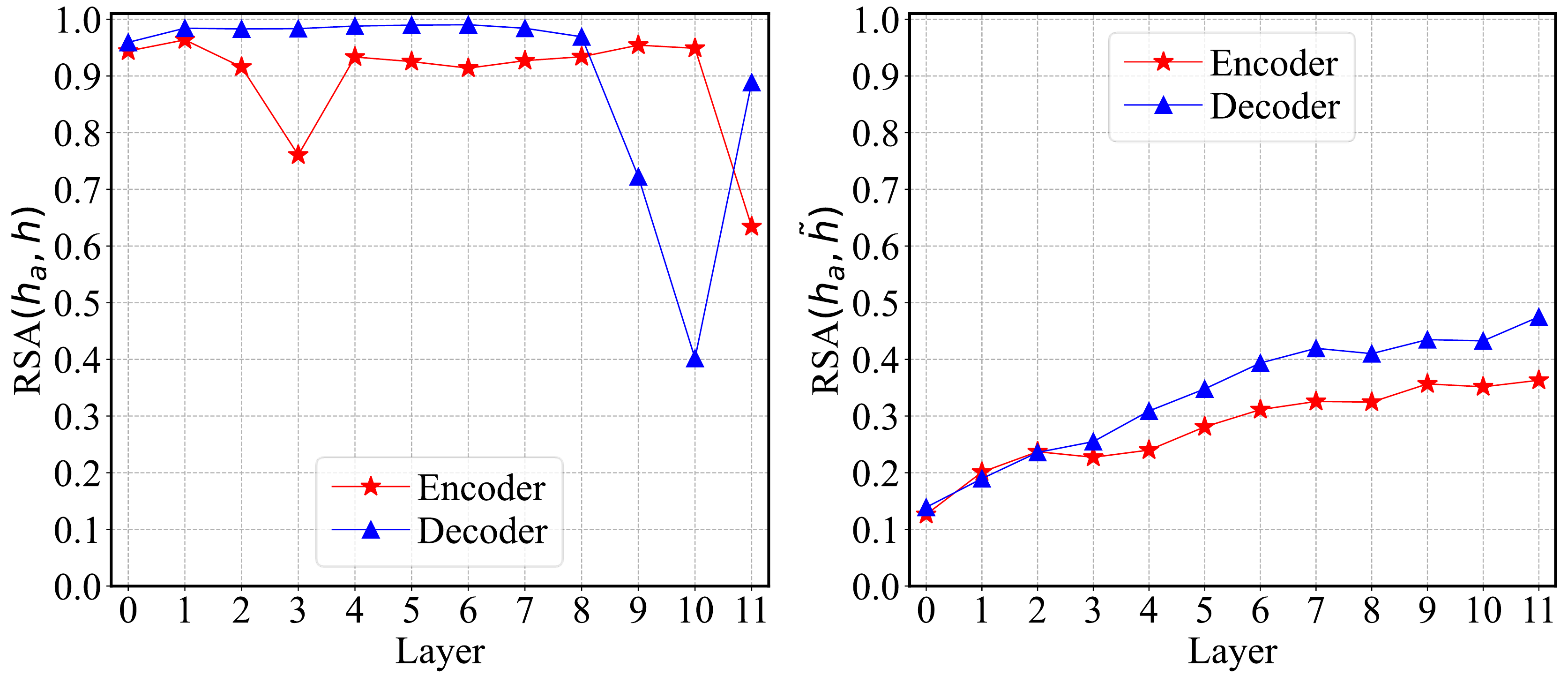}
\vspace{-7mm}
\caption{\textbf{The average RSA similarity across 1k samples} between $\hv_a$ and $\hv$ (left) as well as $\hv_a$ and $\tilde{\hv}$ (right) at each transformer layer. 
The higher the RSA, the more similar (redundant) the representation spaces are. 
}
\label{fig:adapter_rsa}
\vspace{-4mm}
\end{figure}

\pgraph{Redundancy Analysis for Adapter} 
However, recent investigation has shown that some of the information captured by adapters is task-agnostic~\cite{he2021effectiveness}. 
To get the facts, we leverage Representational Similarity Analysis (RSA)~\cite{laakso2000content} to assess the redundancy in representation spaces. 
Specifically, we first tune the pretrained VL-T5~\cite{cho2021unifying} with Pfeiffer~\cite{pfeiffer2020adapterfusion} on 1k samples from VQA v2 training set~\cite{goyal2017making}. 
Then, we randomly sample 1k samples from VQA v2 val set and extract token-level representations (\ie, $\hv$ and $\hv_a$) at each transformer layer as well as the final output representation $\tilde{\hv}$ of transformer encoder/decoder. 
Finally, for each sample, we can obtain $N_t$ token-level representations at each layer, \ie, $\mathbf{H} = \{\hv\}_{i=1}^{N_t} \in \mathbb{R}^{N_t\times d}$, $\mathbf{H}_a = \{\hv_a\}_{i=1}^{N_t}  \in \mathbb{R}^{N_t\times d}$ and $\tilde{\mathbf{H}} = \{\tilde{\hv}\}_{i=1}^{N_t}  \in \mathbb{R}^{N_t\times d}$. 
In each layer, we compute RSA similarity between $\hv_a$ and $\hv$ as well as $\hv_a$ and $\tilde{\hv}$ by 
\begin{align} 
&\text{RSA}(\hv_a, \hv) = f_{\rho}(f_{\text{U}}[\mathbf{H}_a \mathbf{H}_a^{\Transpose}], f_{\text{U}}[\mathbf{H} \mathbf{H}^{\Transpose}]),\\
&\text{RSA}(\hv_a, \tilde{\hv}) = f_{\rho}(f_{\text{U}}[\mathbf{H}_a \mathbf{H}_a^{\Transpose}], f_{\text{U}}[\tilde{\mathbf{H}} \tilde{\mathbf{H}}^{\Transpose}]), 
\label{eq:rsa} 
\end{align}
where, $f_{\text{U}}\left[\cdot\right]$ denotes an operation of taking the upper triangular elements from a matrix, $f_{\rho}$ is a function to compute the Pearson correlation coefficient. 
\Cref{fig:adapter_rsa} illustrates the average RSA similarity across 1k samples, which demonstrates that in transformer layers, the adapter representation $\hv_a$ is redundant with the representation $\hv$ of pretrained VLMs, but has limited correlation to the final output $\tilde{\hv}$.

Intuitively, to transfer pretrained VLMs to downstream tasks efficiently, the representation $\hv_a$ learned by adapter needs to contain as much information as possible from the task-relevant representation $\tilde{\hv}$, while reducing task-irrelevant redundancy with the representation $\hv$ of pretrained VLMs. 
However, \Cref{fig:adapter_rsa} exhibits a counterintuitive result. 
Therefore, in order to improve the effectiveness of adapters, it is crucial to \textit{encourage task-relevant correlation between $\hv_a$ and $\tilde{\hv}$ while reducing task-irrelevant redundancy between $\hv_a$ and $\hv$}.

\subsection{MixPHM Architecture}
\label{sec:mixphm}

As illustrated in \Cref{fig:overall}, MixPHM is also a lightweight module inserted into each transformer block of VLMs. 
We utilize transformer-based encoder-decoder models as underlying pretrained VLMs, which consists of repeated $L$ encoder and $L$ decoder blocks. 
Specifically, for the $l$-th ($1 \le l \le L$) block, we insert a MixPHM composed of a set of $N_e$ PHM-experts $\{E_i^l \}_{i=1}^{N_e}$ after the feed-forward layer to capture the knowledge and learn task-specific information. 
As with the adapter, each PHM-expert is implemented by a bottleneck network with down- and up-projection layers. 

To reduce \textit{parameter redundancy} in MixPHM, we first decompose and reparameterize the projection matrices of experts in MixPHM into low-dimensional subspace. 
Then, we further reduce the number of parameters and transfer information using a strategy of global and local expert weight sharing.
Moreover, a stochastic routing~\cite{zuo2022taming,wang2022adamix} is employed for expert selection to avoid gating networks from introducing additional trainable parameters.

\pgraph{Parameter-Efficient Tuning} 
At each training (tuning) iteration, we randomly select one expert from the inserted $N_e$ PHM-experts in the $l$-th transformer block. Once the expert $E_i^l$ is selected, all inputs in a given batch are processed by the same expert. 
Formally, in the $l$-th block\footnote{For brevity, the superscript $l$ is omitted hereafter.}, for a token input representation $\hv \in \mathbb{R}^d$, the randomly selected $i$-th expert encodes and updates $\hv$ by 
\begin{equation} 
\begin{aligned} 
\hv \leftarrow ~f(\hv \Wmat_i^{\text{dn}})\Wmat_i^{\text{up}} + \hv. 
\end{aligned}
\label{eq:adapter_1} 
\end{equation}
In \cref{eq:adapter_1}, the down-projection matrix $\Wmat_i^{\text{dn}} \in \mathbb{R}^{d\times d_r}$ and up-projection matrix $\Wmat_i^{\text{up}} \in \mathbb{R}^{d_r\times d}$ are firstly decomposed into low-dimensional matrices using PHM, \ie, 
\begin{equation} 
\begin{aligned} 
&\Wmat_i^{\text{dn}} = \sum_{j=1}^{n} \textcolor{Green}{\Smat_{i,j}} \otimes \Amat_{i,j}^{\text{dn}}, 
&\Wmat_i^{\text{up}} = \sum_{j=1}^{n} \textcolor{Green}{\Smat_{i,j}} \otimes \Amat_{i,j}^{\text{up}}, 
\label{eq:ada_phm}
\end{aligned}
\end{equation}
where, $\textcolor{Green}{\Smat_{i,j}} \in \mathbb{R}^{n\times n}$, $\Amat_{i,j}^{\text{dn}} \in \mathbb{R}^{\frac{d}{n} \times \frac{d_r}{n}}$, $\Amat_{i,j}^{\text{up}} \in \mathbb{R}^{\frac{d_r}{n} \times \frac{d}{n}}$. 
To be more parameter-efficient, the matrix $\Amat_{i,j}^{\text{dn}}$ ($\Amat_{i,j}^{\text{up}}$) is further factorized into two low-rank matrices by
\begin{equation} 
\begin{aligned} 
&\Amat_{i,j}^{\text{dn}} = \Tmat_{i,j}^{\text{dn}} (\Umat_{i,j}^{\text{dn}})^{\Transpose}
, 
&\Amat_{i,j}^{\text{up}} = \Tmat_{i,j}^{\text{up}} (\Umat_{i,j}^{\text{up}})^{\Transpose}
, 
\label{eq:ada_phm_lr}
\end{aligned}
\end{equation}
where, $\Tmat_{i,j}^{\text{dn}} \in \mathbb{R}^{\frac{d}{n}\times d_k}$, $\Umat_{i,j}^{\text{dn}} \in \mathbb{R}^{\frac{d_r}{n}\times d_k}$, $\Tmat_{i,j}^{\text{up}} \in \mathbb{R}^{\frac{d_r}{n}\times d_k}$, $\Umat_{i,j}^{\text{up}} \in \mathbb{R}^{\frac{d}{n}\times d_k}$, and $d_r$ is the rank of these matrices. 
Finally, we learn the weight matrices of the $i$-th PHM-expert by
\begin{align} 
&\Wmat_i^{\text{dn}} = \sum_{j=1}^{n} \textcolor{Green}{\Smat_{i,j}} \otimes (\Tmat_{i,j}^{\text{dn}} (\Umat_{i,j}^{\text{dn}})^{\Transpose}), 
\\
&\Wmat_i^{\text{up}} = \sum_{j=1}^{n} \textcolor{Green}{\Smat_{i,j}} \otimes (\Tmat_{i,j}^{\text{up}} (\Umat_{i,j}^{\text{up}})^{\Transpose}). 
\label{eq:ada_lphm}
\end{align}

\pgraph{Information Sharing across PHM-Experts} 
When tuning pretrained VLMs with MixPHM on a downstream dataset, the set of $n$ matrices $\{\textcolor{Green}{\Smat_{i,j}}\in \mathbb{R}^{n\times n}\}_{j=1}^n$ of the $i$-th PHM-expert are globally shared among all PHM-experts across transformer blocks to capture general information for the target task. 
In other words, the $\textcolor{Green}{\Smat} = \{\textcolor{Green}{\Smat}_j\}_{j=1}^n \in \mathbb{R}^{n\times n\times n}$ of one expert is shared between all added MixPHMs. 
On the contrary, $\{\Amat_{i,j}^{\text{dn}}\}_{j=1}^n$ and $\{\Amat_{i,j}^{\text{up}}\}_{j=1}^n$ are expert-specific weight matrices which are unique to each PHM-expert. 
To better transfer information between PHM-experts of MixPHM and further reduce parameter redundancy, we locally share $\{\Amat_{i,j}^{\text{dn}}\}_{j=1}^n$ among PHM-experts in each MixPHM.

At this point, the total number of trainable parameters inserted into pretrained VLMs using MixPHM is reduced from the original $4LN_e(dd_r)$ to $2Ld_k(d+d_r)(N_e + 1) + n^3$.

\subsection{Redundancy Regularization} 
\label{sec:L_ra}

Motivated by the insight discussed in \cref{sec:revist}, we propose redundancy regularization. 
Specifically, for the MixPHM in the $l$-th transformer block, we ensemble its token-level output representation $\{\hv_a\}_{i=1}^{N}$ and its residual $\{\hv\}_{i=1}^{N}$ of a batch to $\Zmat_a \in \mathbb{R}^{N \times d}$ and $\Zmat \in \mathbb{R}^{N \times d}$, $N=N_bN_t$ ($N_b$ indicates batch size). 
For the transformer encoder/decoder, we average the final output representation $\{\tilde{\hv}\}_{i=1}^{N}$ of a batch along the token dimension and obtain a global task-relevant representation $\{\bar{\hv} \}_{i=1}^{N_b}$. 
Then, the redundancy regularization can be expressed by
\begin{equation}
\mathcal{L}_{\text{Ra}} \triangleq \sum_i^N \sum_{j \neq i}^N \frac{\Zmat_a \Zmat^{\Transpose}}{\left\|\Zmat_a\right\|_2 \left\|\Zmat\right\|_2} - \sum_i^{N_b}\sum_j^{N_t} \hat{\mathcal{I}}({\hv_{a}}_{i,j}; \bar{\hv}_{i}),
\label{eq:L_Ra} 
\end{equation} 
where, $\left\|\cdot\right\|_2$ denotes the $L_2$ norm, ${\hv_{a}}_{i,j}$ is the output representation of the $j$-th token of the $i$-th sample in a batch, and $\hat{\mathcal{I}}(\cdot; \cdot)$ means an estimation of mutual information (MI), which is used to maximize the correction between two representations and computed by the JSD MI estimator~\cite{hjelm2019learning}. 
In redundancy regularization $\mathcal{L}_{\text{Ra}}$, the first term is a redundancy reduction term, which encourages $\hv_a$ to discard task-irrelevant information from pretrained VLMs by encouraging the off-diagonal elements of the cosine similarity matrix between $\hv_a$ and $\hv$ to zero. 
The second term aims to advocate $\hv_a$ contain more task-relevant information from downstream datasets by maximizing the MI between $\hv_a$ and $\bar{\hv}$.

Formulating VQA as a generative task, the training objective is to minimize the negative log-likelihood of answer $y$ tokens given input image $I$ and question $Q$.  
Therefore, the total training loss in parameter-efficient tuning is 
\begin{equation}
\begin{aligned}
\mathcal{L} = -\sum_{j=1}^{|y|}\log P_{\thetav}(y_j|y_{<j}; I, Q) + \alpha \mathcal{L}_{\text{Ra}}, 
\label{eq:L_total}
\end{aligned}
\end{equation} 
where, $\alpha$ is a factor to balance redundancy regularization.

\subsection{Inference} 
\label{sec:inference}

In contrast to the stochastic routing utilized during training, we adopt a weight aggregation strategy~\cite{wortsman2022model} to obtain a final PHM-expert for each MixPHM during inference. 
Specifically, one MixPHM has $N_e$ PHM-experts. 
When learning weights in a low-rank subspace, each expert has $2n$ expert-specific matrices $\{\Tmat_j^{\text{up}}, \Umat_j^{\text{up}}\}_{j=1}^n$, and the $N_e$ experts have the same $2n$ locally-shared matrices $\{\Tmat_j^{\text{dn}}, \Umat_j^{\text{dn}}\}_{j=1}^n$ as well as $n$ globally-shared matrices $\{ \textcolor{Green}{\Smat_j} \}_{j=1}^n$. 
To obtain weights of the final PHM-expert, we first merge the weights of up-projection matrices by averaging the corresponding $N_e$ weight matrices. 
Mathematically, the $j$-th up-projection matrices can be computed with 
\begin{equation}
\begin{aligned}
\widetilde{\Tmat}_j^{\text{up}} = \frac{1}{N_e} \sum_{i=1}^{N_e} \Tmat_{ji}^{\text{up}}, \quad
\widetilde{\Umat}_j^{\text{up}} = \frac{1}{N_e} \sum_{i=1}^{N_e} \Umat_{ji}^{\text{up}}. 
\end{aligned}
\label{eq:weight_merge}
\end{equation} 
Due to the global and local weight sharing, we need not perform weight aggregation on $\textcolor{Green}{\Smat}$ and $\{\Tmat_j^{\text{dn}}, \Umat_j^{\text{dn}}\}_{j=1}^n$. 
Finally, we employ the merged expert to compute the output representations of MixPHM at each transformer block.

%% file: section/exp.tex
\input{table/t1_pet_base.tex}

\section{Experiment}

\subsection{Experimental Setting}
\label{sec:exp_setting}

\pgraph{Datasets and Metrics} We conduct experiments on three datasets, VQA v2~\cite{goyal2017making}, GQA~\cite{hudson2019gqa}, and OK-VQA~\cite{marino2019ok}. 
To simulate the low-resource setting for VQA, we follow the work~\cite{chen2022revisiting} and consider the training data size $N_{\mathcal{D}}$ for low-resource VQA to be smaller than 1,000. 
For more practical low-resource learning, we follow \textit{true few-shot learning}~\cite{perez2021true,gao2021making} and utilize the development set $\mathcal{D}_{\text{dev}}$, which has the same size with the training set $\mathcal{D}_{\text{train}}$ (\ie, $|\mathcal{D}_{\text{dev}}|=$  $|\mathcal{D}_{\text{train}}|=$ $N_{\mathcal{D}}$), instead of using large-scale validation set, for best model selection and hyperparameter tuning. 
Specifically, we conduct experiments on $N_{\mathcal{D}} \in \{16, 32, 64, 100, 500, 1000\}$. 
To construct the $\mathcal{D}_{\text{train}}$ and $\mathcal{D}_{\text{dev}}$ of VQA v2, we randomly sample $2N_{\mathcal{D}}$ samples from its training set and divide them equally into the $\mathcal{D}_{\text{train}}$ and $\mathcal{D}_{\text{dev}}$. 
Analogously, we construct $\mathcal{D}_{\text{train}}$ and $\mathcal{D}_{\text{dev}}$ for GQA and OK-VQA. 
VQA-Score~\cite{antol2015vqa} is the accuracy metric of the low-resource VQA task.

\pgraph{Baselines} We compare our method with several state-of-the-art parameter-efficient tuning methods and finetuning. 
For a fair comparison, we perform hyperparameter search on their key hyperparameters (KHP) and report their best performance. 
Specifically, 
\begin{itemize}[leftmargin=9pt,itemsep=1pt,topsep=1pt,parsep=0pt]
\item \textbf{BitFit}~\cite{zaken2022bitfit} only tunes the bias weights of pretrained models while keeping the rest parameters frozen. 
\item \textbf{LoRA}~\cite{hu2022lora} tunes additional low-rank matrices, which are used to approximate the query and value weights in each transformer self-attention and cross-attention layer. 
KHP is the matrix rank ($r$). 
\item \textbf{Compacter}~\cite{karimi2021compacter} adds adapters after each feed-forward layer of transformer blocks and reparameterizes adapter weights with low-rank PHM layers~\cite{zhang2021beyond}. 
KHP are the number of summations of Kronecker product ($n$), the bottleneck dimension ($d_r$), and $r$. 
\item \textbf{Houlsby}~\cite{houlsby2019parameter} adds adapters after self-attention and feed-forward layers in each transformer block. KHP is $d_r$. 
\item \textbf{Pfeiffer}~\cite{pfeiffer2020adapterfusion} is to determine the location of adapter based on pilot experiments. In this work, we place it after each transformer feed-forward layer. KHP is $d_r$. 
\item \textbf{AdaMix}~\cite{wang2022adamix} adds multiple adapters after each transformer feed-forward layer in a MoE manner. 
KHP are the number of adapters ($N_e$), and $d_r$. 
\end{itemize}

\input{table/t2_fs_vqa.tex}

\pgraph{Implementation Details} 
We use four pretrained VLMs, \ie, VL-T5~\cite{cho2021unifying}, X-VLM~\cite{zeng2022multi}, BLIP~\cite{li2022blip}, and OFA$_{\text{Base}}$~\cite{wang2022ofa}, as underlying encoder-decoder transformers, which formulate VQA as a generation task in finetuning and thus do not introduce additional parameters from VQA heads. 
Since the original pretraining datasets employed by VL-T5 contain samples of the above VQA datasets, we instead load the weights\footnote{\href{https://github.com/woojeongjin/FewVLM}{https://github.com/woojeongjin/FewVLM}} released by Jin~\etal~\cite{jin2022good}, which is re-trained without the overlapped samples. 
All results are reported across five seeds $\{13, 21, 42, 87, 100\}$. 
More details and hyperparameter setups are provided in \cref{sec:imp_detail}.

\input{table/t3_abl_regularizer.tex}
\subsection{Low-Resource Visual Question Answering}

\Cref{tab:vqa_pet_base} shows the results with pretrained VL-T5~\cite{cho2021unifying} on three datasets. 
Overall, our MixPHM outperforms state-of-the-art parameter-efficient tuning methods and is the only one that consistently outperforms full finetuning. 
Next, we elaborate on the different comparisons.

\pgraph{Comparison with AdaMix} 
AdaMix and MixPHM adopt MoE to boost the capacity of adapters, but MixPHM considers further reducing parameter redundancy and learning more task-relevant representations. 
Table~\ref{tab:vqa_pet_base} shows that on all datasets with different $N_{\mathcal{D}}$, our MixPHM markedly outperforms AdaMix and full finetuning, while AdaMix fails to outperform full finetuning (except VQA v2 with $N_{\mathcal{D}}=32$). 
This result demonstrates the effectiveness of MixPHM in terms of performance and parameter efficiency, which also suggests the importance of prompting task-relevant correction while reducing parameter redundancy.

\pgraph{Comparison with Houlsby and Pfeiffer} 
PHM-expert in MixPHM has the same bottleneck structure with adapter. 
However, PHM-expert is more parameter-efficient due to the reduction of parameter redundancy and can better capture task-relevant information owing to the proposed redundancy regularization. 
The result in Table~\ref{tab:vqa_pet_base} shows that compared to Finetuning, the performance of Houlsby and Pfeiffer falls far short of the ceiling performance in most low-resource settings. 
Conversely, the proposed MixPHM exhibits advantages in terms of performance and parameter efficiency under all dataset settings.

\pgraph{Comparison with Compacter} 
To reduce parameter redundancy, Compacter and MixPHM reparameterize adapter weights with low-rank PHM. 
However, in reducing parameter redundancy, MixPHM encourages task-relevant correction with redundancy regularization and improves the model capacity with MoE, avoiding overfitting and under-parameterization concerns. 
Table~\ref{tab:vqa_pet_base} shows that Compacter does not perform as expected. 
One possible explanation is that Compacter is under-parameterized on the low-resource VQA task. 
Because too few trainable parameters do not guarantee that model capture enough task-relevant information in the tuning stage. 
This suggests that when tuning pretrained VLMs for low-resource VQA, it is necessary to balance the effectiveness and parameter efficiency.

\pgraph{Comparison with LoRA and BitFit} 
LoRA and BitFit are two typical parameter-efficient methods that tune a part of parameters of original pretrained VLMs and are more parameter-efficient. 
The results are shown in Table~\ref{tab:vqa_pet_base}. 
We observe that compared with MixPHM, the tunable parameters of LoRA and BitFit are relatively lightweight. 
However, their performance trails much below MixPHM. 
In particular, the performance gap becomes larger as $N_{\mathcal{D}}$ increases. 
Similar to the discussion on Compacter, too few trainable parameters in the tuning process may lead to overfitting on the given dataset.

\pgraph{Comparison with SoTA Few-Shot Learner} 
Few-shot VQA is a special case of low-resource VQA. 
The comparisons with SoTA multimodal few-shot learner are shown in Table~\ref{tab:vqa_few_shot}. 
Frozen~\cite{tsimpoukelli2021multimodal} and PICa~\cite{yang2022empirical} are two in-context learning methods that adopt prompt-tuning to transfer large-scale language models (\ie, GPT-2 and GPT-3~\cite{brown2020language}) without parameter tuning. 
While FewVLM~\cite{jin2022good} is a prompt-based full finetuning method to adapt VL-T5, which inserts hand-crafted prompts into model inputs and finetunes full model parameters. 
We utilize MixPHM to tune FewVLM in a parameter-efficient manner. 
With only a tiny number of parameters tuned, MixPHM outperforms FewVLM in few-shot performance, especially on OK-VQA (\textbf{19.2} vs. 15.0). 
This demonstrates the superiority of MixPHM in terms of performance and parameter efficiency.

\input{table/t4_abl_param_redundancy.tex}

\subsection{Ablation Study} 

We conduct all ablated experiments with pretrained VL-T5 on VQA v2, GQA, and OK-VQA with $N_{\mathcal{D}}=64$.

\pgraph{Effectiveness of Redundancy Regularization} 
To demonstrate the effectiveness of the proposed redundancy regularization, we first introduce a consistency regularizer $\mathcal{L}_{\text{cs}}$~\cite{zuo2022taming} for comparison. 
Moreover, to further analyze the contribution of different terms in $\mathcal{L}_{\text{Ra}}$, we consider two variations of $\mathcal{L}_{\text{Ra}}$: (\emph{i})
${\mathcal{L}_{\text{Ra}}^{\text{I}}}$: only using the first term in \cref{eq:L_Ra} as the regularizer during training. 
(\emph{ii}) ${\mathcal{L}_{\text{Ra}}^{\text{II}}}$: only using the second term in \cref{eq:L_Ra} during training. 
\Cref{tab:abl_ra} shows that $\mathcal{L}_{\text{cs}}$ improves MixPHM performance only on GQA and the improvement is minor. 
In contrast, $\mathcal{L}_{\text{Ra}}$ shows a significant improvement in MixPHM performance on all datasets. 
This observation demonstrates the effectiveness and superiority of the proposed regularizer $\mathcal{L}_{\text{Ra}}$. 
Furthermore, analyzing the impact of ${\mathcal{L}^{\text{I}}_{\text{Ra}}}$ and ${\mathcal{L}^{\text{II}}_{\text{Ra}}}$ on MixPHM performance, we find that only reducing the redundancy between the representation of MixPHM and the representation of pretrained VLMs (\ie, ${\mathcal{L}^{\text{I}}_{\text{Ra}}}$) makes limited contribution to performance gains. 
However, the joint effect of ${\mathcal{L}^{\text{I}}_{\text{Ra}}}$ and ${\mathcal{L}^{\text{II}}_{\text{Ra}}}$ is better than ${\mathcal{L}^{\text{II}}_{\text{Ra}}}$ alone. 
This suggests that the trade-off between reducing task-irrelevant redundancy and prompting task-relevant correlation is critical for MixPHM.

\pgraph{Impact of Reducing Parameter Redundancy} 
MixPHM reduces parameter redundancy by first decomposing expert weights with PHM (D1) and then reparameterizing the decomposed weights (D2). 
We ablate D1 and D2 to analyze their effects on MixPHM performance (\ie, the third column in \Cref{tab:abl_param_redundancy}). 
In addition, weight sharing can further reduce parameter redundancy in MixPHM. 
We thus conduct ablation on different meaningful combinations of shared weights in the fourth column of \Cref{tab:abl_param_redundancy}. 
Aside from the globally shared matrices ($\textcolor{Green}{\Smat}$), we also locally share down-projection ($\Amat^{\text{dn}}$) or up-projection ($\Amat^{\text{up}}$) matrices between experts in one MixPHM. 
\Cref{tab:abl_param_redundancy} shows that there is a trade-off between parameter efficiency and performance, \ie, excessive parameter reduction may harm performance. 
Therefore, we advocate reducing parameter redundancy while maintaining model capacity.

\pgraph{Impact of Hyperparameters} 
Results on hyperparameters ($N_e$, $d_r$, $d_k$, $n$, $\alpha$) are available in \cref{sec:imp_abl}.

\subsection{Discussion}

\begin{figure}[!t]
\centering
\includegraphics[width=1.0\linewidth]{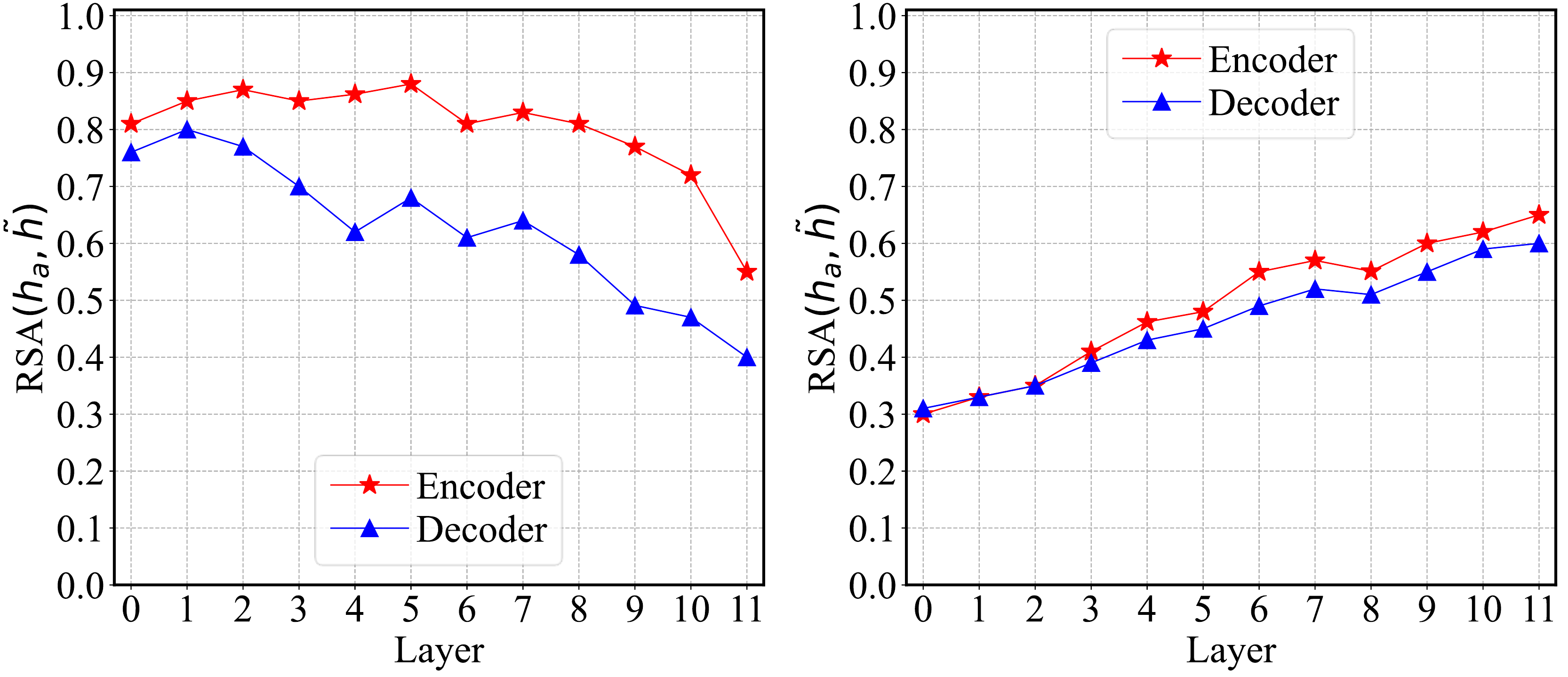}
\vspace{-7mm}
\caption{\textbf{The average RSA similarity of MixPHM} between $\hv_a$ and $\hv$ (left) as well as $\hv_a$ and $\tilde{\hv}$ (right) at each transformer layer. 
}
\label{fig:mixphm_rsa}
\vspace{-3mm}
\end{figure}

\pgraph{Redundancy Analysis of MixPHM} 
In this paper, we propose an insight that improving the effectiveness of adapters via reducing task-irrelevant redundancy and promoting task-relevant correlation in representations. 
To assess whether our method actually leads to performance improvements based on this insight, we conduct the redundancy analysis in the MixPHM representation space under the same experimental settings as described in \cref{sec:revist}. 
\Cref{fig:mixphm_rsa} illustrates the RSA similarity across 1k samples on VQA v2. 
Compared with the redundancy analysis result of adapter shown in \Cref{fig:adapter_acc}, we observe that MixPHM markedly reduces the representation redundancy between $\hv_a$ and $\hv$, while increasing the representation correlation between $\hv_a$ and $\tilde{\hv}$. 
This finding provides a perceptive demonstration for the soundness of our motivation and the effectiveness of our method.

\input{table/t5_dis_generalization.tex}

\pgraph{Generalizability to Other Pretrained VLMs}
To demonstrate the generalization capability of our method on other pretrained VLMs, we apply MixPHM to adapt pretrained X-VLM~\cite{zeng2022multi}, BLIP~\cite{li2022blip}, and OFA$_{\text{Base}}$~\cite{wang2022ofa} for the low-resource VQA task. 
\Cref{tab:generalization} presents a lite comparison between our method and full finetuning on VQA v2. 
We observe that MixPHM consistently outperforms full finetuning in all settings, with significant performance gains observed when $N_{\mathcal{D}}$ $\in$ $\{500,1000\}$. 
Notably, the largest performance gaps from finetuning are achieved by X-VLM (\textbf{+4.14}), BLIP (\textbf{+4.80}), and OFA$_{\text{Base}}$ (\textbf{+3.20}) at $N_{\mathcal{D}}$=1000, $N_{\mathcal{D}}$=500, and $N_{\mathcal{D}}$=1000, respectively. 
These findings demonstrate the generalizability of MixPHM to various pretrained VLMs. 
More comparisons between parameter-efficient methods are available in \cref{sec:imp_xvlm}. 

%% file: table/t1_pet_base.tex
\begin{table*}[!t]
\centering
\footnotesize  
\setlength{\tabcolsep}{2.49mm}{
\begin{tabularx}{\linewidth}{@{}llrrrrrrrr}
\toprule

\multirow{2}{*}{Dataset} 
&\multirow{2}{*}{Method} 
&\multicolumn{2}{c}{\#Param} 
&\multicolumn{6}{c}{\#Sample}
\\

& &\multicolumn{1}{c}{(M)} 
&\multicolumn{1}{c}{(\%)}

&$N_{\mathcal{D}}$=16 
&$N_{\mathcal{D}}$=32 
&$N_{\mathcal{D}}$=64 
&$N_{\mathcal{D}}$=100 
&$N_{\mathcal{D}}$=500 
&$N_{\mathcal{D}}$=1,000 
\\ 

\midrule
\rowcolor{gray!30}
\cellcolor{white}
&Finetuning &224.54 &100\%
&41.82$~\std{1.58}$
&43.09$~\std{3.10}$
&46.87$~\std{0.57}$
&48.12$~\std{0.87}$
&53.46$~\std{0.41}$
&55.56$~\std{0.13}$
\\ 

&BitFit~\cite{zaken2022bitfit} 
&0.29  
&0.13\%
&40.61$~\std{4.15}$ 
&43.86$~\std{2.19}$
&46.14$~\std{1.00}$
&47.53$~\std{0.67}$
&51.91$~\std{0.40}$
&53.18$~\std{0.58}$
\\

&LoRA~\cite{hu2022lora} 
&0.44 
&0.20\%
&41.60$~\std{2.27}$
&42.62$~\std{2.41}$
&45.36$~\std{1.66}$
&47.57$~\std{0.91}$
&51.93$~\std{0.38}$
&54.15$~\std{0.45}$
\\ 

&Compacter~\cite{karimi2021compacter} 
&0.34 
&0.15\%
&39.28$~\std{1.87}$ 
&42.47$~\std{2.76}$
&44.91$~\std{1.27}$
&46.28$~\std{1.37}$
&51.21$~\std{0.90}$
&53.39$~\std{0.54}$
\\ 

&Houlsby~\cite{houlsby2019parameter} 
&4.76 
&2.12\%
&\text{\color{blue}{41.71$~\std{2.16}$}} 
&44.01$~\std{2.09}$
&45.11$~\std{1.40}$
&\text{\color{blue}{47.71$~\std{0.78}$}}
&52.27$~\std{1.05}$
&\text{\color{blue}{54.31$~\std{0.34}$}}
\\ 

&Pfeiffer~\cite{pfeiffer2020adapterfusion} 
&2.38 
&1.06\%
&41.48$~\std{1.86}$ 
&\text{\color{blue}{44.18$~\std{2.13}$}}
&45.93$~\std{1.11}$
&47.42$~\std{1.15}$
&\text{\color{blue}{52.35$~\std{0.52}$}}
&53.98$~\std{0.38}$
\\

&AdaMix~\cite{wang2022adamix} 
&5.92 
&2.64\%
&40.59$~\std{2.05}$
&43.42$~\std{2.08}$
&\text{\color{blue}{46.70$~\std{1.32}$}} 
&47.34$~\std{0.91}$
&51.72$~\std{1.05}$
&54.12$~\std{0.63}$
\\

\multirow{-8}{*}{\makecell[l]{VQA v2\\\cite{goyal2017making}}}
&\text{MixPHM} 
&0.87
&0.39\%
&\text{\color{red}{43.13$~\std{1.78}$}}
&\text{\color{red}{45.97$~\std{2.01}$}}
&\text{\color{red}{48.26$~\std{0.56}$}}
&\text{\color{red}{49.91$~\std{0.76}$}}
&\text{\color{red}{54.30$~\std{0.33}$}}
&\text{\color{red}{56.11$~\std{0.40}$}}
\\ 

\midrule[0.5pt]
\rowcolor{gray!30}
\cellcolor{white}
&Finetuning &224.54 &100\%
&28.24$~\std{2.08}$ 
&30.80$~\std{2.49}$
&34.22$~\std{0.59}$
&36.15$~\std{0.99}$ 
&41.49$~\std{0.54}$ 
&43.04$~\std{0.57}$ 
\\ 

&BitFit~\cite{zaken2022bitfit} 
&0.29 
&0.13\%
&26.13$~\std{2.83}$ 
&29.00$~\std{4.81}$ 
&34.25$~\std{1.16}$
&35.91$~\std{1.22}$
&40.08$~\std{0.42}$
&41.84$~\std{0.15}$
\\

&LoRA~\cite{hu2022lora} 
&0.44 
&0.20\%
&\text{\color{blue}{26.89$~\std{2.74}$}} 
&\text{\color{blue}{30.40$~\std{2.27}$}}
&\text{\color{blue}{34.40$~\std{0.99}$}}
&\text{\color{blue}{36.14$~\std{1.10}$}}
&40.20$~\std{1.02}$
&\text{\color{blue}{42.06$~\std{1.12}$}}
\\ 

&Compacter~\cite{karimi2021compacter} 
&0.34 
&0.15\%
&23.70$~\std{2.10}$
&27.18$~\std{2.61}$ 
&32.70$~\std{1.30}$
&35.28$~\std{1.45}$
&38.68$~\std{0.50}$
&41.17$~\std{0.95}$
\\

&Houlsby~\cite{houlsby2019parameter} 
&4.76 
&2.12\%
&25.13$~\std{2.32}$ 
&28.34$~\std{1.17}$
&33.23$~\std{0.94}$
&35.88$~\std{1.79}$
&\text{\color{blue}{40.85$~\std{0.48}$}}
&41.90$~\std{0.72}$
\\ 

&Pfeiffer~\cite{pfeiffer2020adapterfusion} 
&2.38 
&1.06\%
&25.08$~\std{1.81}$ 
&29.18$~\std{1.32}$ 
&32.97$~\std{0.84}$
&35.08$~\std{1.01}$
&40.30$~\std{0.40}$
&41.39$~\std{0.27}$
\\ 

&AdaMix~\cite{wang2022adamix} 
&5.92 
&2.64\%
&24.62$~\std{2.34}$ 
&28.01$~\std{1.33}$
&32.74$~\std{0.96}$
&35.64$~\std{0.94}$
&40.14$~\std{0.42}$
&41.97$~\std{0.86}$
\\ 

\multirow{-8}{*}{\makecell[l]{GQA\\\cite{hudson2019gqa}}}
&MixPHM
&0.87
&0.39\%
&\text{\color{red}{28.33$~\std{2.63}$}}
&\text{\color{red}{32.40$~\std{2.52}$}}
&\text{\color{red}{36.75$~\std{0.55}$}}
&\text{\color{red}{37.40$~\std{0.87}$}}
&\text{\color{red}{41.92$~\std{0.55}$}}
&\text{\color{red}{43.81$~\std{0.50}$}}
\\ 

\midrule[0.5pt]
\rowcolor{gray!30}
\cellcolor{white}
&Finetuning &224.54 &100\%
&11.66$~\std{2.08}$
&14.20$~\std{0.78}$
&16.65$~\std{1.02}$
&18.28$~\std{0.67}$
&24.07$~\std{0.40}$
&26.66$~\std{0.72}$
\\

&BitFit~\cite{zaken2022bitfit} 
&0.29
&0.13\%
&\text{\color{blue}{11.29$~\std{1.79}$}} 
&\text{\color{blue}{13.66$~\std{1.49}$}}
&15.29$~\std{0.57}$
&16.51$~\std{0.53}$
&22.54$~\std{0.57}$
&24.80$~\std{0.63}$
\\

&LoRA~\cite{hu2022lora} 
&0.44
&0.20\%
&10.26$~\std{1.53}$
&12.46$~\std{1.82}$
&\text{\color{blue}{15.95$~\std{0.38}$}}
&\text{\color{blue}{17.03$~\std{0.82}$}}
&23.02$~\std{0.41}$
&25.26$~\std{0.53}$

\\ 

&Compacter~\cite{karimi2021compacter} 
&0.34 
&0.15\%
&9.64$~\std{2.73}$ 
&11.04$~\std{1.39}$
&13.57$~\std{1.07}$
&15.92$~\std{1.18}$
&22.20$~\std{0.89}$
&24.52$~\std{0.59}$
\\

&Houlsby~\cite{houlsby2019parameter}
&4.76 
&2.12\%
&9.79$~\std{1.71}$
&12.25$~\std{2.13}$
&15.04$~\std{1.25}$
&16.58$~\std{0.65}$
&22.67$~\std{0.77}$
&25.04$~\std{0.44}$
\\ 

&Pfeiffer~\cite{pfeiffer2020adapterfusion} 
&2.38
&1.06\%
&9.06$~\std{0.53}$
&11.39$~\std{0.79}$
&14.23$~\std{1.54}$
&16.34$~\std{0.79}$
&22.90$~\std{1.03}$
&\text{\color{blue}{26.70$~\std{0.71}$}}
\\ 

&AdaMix~\cite{wang2022adamix} 
&5.92 
&2.64\%
&8.39$~\std{1.20}$
&11.55$~\std{1.37}$
&13.66$~\std{2.29}$
&16.27$~\std{0.92}$
&\text{\color{blue}{23.20$~\std{0.78}$}}
&26.34$~\std{0.88}$
\\ 

\multirow{-8}{*}{\makecell[l]{OK-VQA\\\cite{marino2019ok}}}
&MixPHM
&0.87
&0.39\%

&\text{\color{red}{13.87$~\std{2.39}$}} 
&\text{\color{red}{16.03$~\std{1.23}$}}
&\text{\color{red}{18.58$~\std{1.42}$}} 
&\text{\color{red}{20.16$~\std{0.97}$}}
&\text{\color{red}{26.08$~\std{0.88}$}} 
&\text{\color{red}{28.53$~\std{0.85}$}} 
\\ 

\bottomrule
\end{tabularx}
}
\vspace{-3mm}
\caption{
\textbf{Experimental results with pretrained VL-T5.} 
The average VQA-Score with standard deviation across 5 seeds are evaluated on VQA v2 validation set, GQA test-dev, and OK-VQA test set. 
The \text{\color{red}{best}} and \text{\color{blue}{second best}} parameter-efficient tuning methods are highlighted. 
The number of tuned parameters and the percentage of tuned parameters relative to VL-T5 (\ie, 224.54M) are reported. 
}
\label{tab:vqa_pet_base}
\vspace{-3mm}
\end{table*}

%% file: table/t2_fs_vqa.tex
\begin{table}[!t]
\footnotesize
\setlength{\tabcolsep}{1.3mm}{
\begin{tabularx}{\linewidth}{@{}lccccc@{}}
\toprule
\multirow{1}{*}{Method} 
&Model Size 
&\#Param (\%)
&\multirow{1}{*}{VQAv2}
&\multirow{1}{*}{GQA} 
&\multirow{1}{*}{OK-VQA} 
\\

\hline

Frozen~\cite{tsimpoukelli2021multimodal} 
&7B
&-
&38.2 &12.6 & -  
\\ 

PICa-Base~\cite{yang2022empirical} 
&175B 
&-
&\text{54.3} &\text{43.3} &- 
\\

PICa-Full~\cite{yang2022empirical} 
&175B 
&-
&\textbf{56.1} &\textbf{48.0} &- 
\\
FewVLM~\cite{jin2022good}  
&225M  
&100\% 
&48.2 
&32.2 
&15.0 
\\ 

MixPHM$^{\dagger}$
&226M
&0.39\%
&49.3
&33.4
&\textbf{19.2}
\\

\bottomrule
\end{tabularx}
}
\vspace{-3mm}
\caption{
\textbf{Comparison with few-shot learner ($N_{\mathcal{D}}$=64).} 
FewVLM is a prompt-based full finetuning method. 
MixPHM$^{\dagger}$ means using MixPHM to tune FewVLM in a parameter-efficient manner. 
}
\label{tab:vqa_few_shot}
\vspace{-5mm}
\end{table}

%% file: table/t3_abl_regularizer.tex
\begin{table}[!t]
\footnotesize 
\centering 
\setlength{\tabcolsep}{3.67mm}{
\begin{tabularx}{\linewidth}{lcccc@{}}
\toprule
\multirow{1}{*}{Method} 
&\multicolumn{1}{c}{VQA v2}
&\multicolumn{1}{c}{GQA}
&\multicolumn{1}{c}{OK-VQA}
\\ 
\midrule

\rowcolor{gray!20}
Finetuning
&46.87$~\std{0.57}$
&34.22$~\std{0.59}$
&16.65$~\std{1.02}$
\\

\multicolumn{1}{c}{MixPHM$^*$}
&47.30$~\std{0.67}$ 
&34.66$~\std{0.78}$ 
&18.05$~\std{1.16}$ 
\\ 

+ ${\mathcal{L}_{\text{cs}}}$
&46.70$~\std{0.66}$ 
&34.83$~\std{1.35}$ 
&17.37$~\std{1.38}$ 
\\

+ ${\mathcal{L}^{\text{I}}_{\text{Ra}}}$
&47.42$~\std{0.71}$ 
&34.69$~\std{0.96}$ 
&18.25$~\std{1.54}$ 
\\

+ ${\mathcal{L}^{\text{II}}_{\text{Ra}}}$
&47.71$~\std{0.85}$ 
&36.10$~\std{0.83}$ 
&18.21$~\std{1.08}$
\\

\rowcolor{DeepSkyBlue!20}
+ $\mathcal{L}_{\text{Ra}}$
&\textbf{48.26$~\std{0.56}$}
&\textbf{36.75$~\std{0.55}$}
&\textbf{18.58$~\std{1.42}$}
\\ 

\bottomrule
\end{tabularx}
}
\vspace{-3mm}
\caption{\textbf{Ablation on different regularizers with $N_{\mathcal{D}}=64$.} 
MixPHM$^*$ means the baseline without any regularizer. 
${\mathcal{L}_{\text{cs}}}$ is a consistency regularizer~\cite{zuo2022taming}. 
${\mathcal{L}^{\text{I}}_{\text{Ra}}}$ and ${\mathcal{L}^{\text{II}}_{\text{Ra}}}$ indicate that only using the first and the second term of $\mathcal{L}_{\text{Ra}}$, respectively. 
}
\label{tab:abl_ra}
\vspace{-3mm}
\end{table}

%% file: table/t4_abl_param_redundancy.tex
\begin{table}[!t]
\centering
\footnotesize
\setlength{\tabcolsep}{0.8mm}{
\begin{tabularx}{\linewidth}{@{}cc|ccc|c|ccc@{}}
\hline

\multicolumn{2}{c|}{WD} 
&\multicolumn{3}{c|}{WS} 
&\#Param 
&\multirow{2}{*}{VQA v2}
&\multirow{2}{*}{GQA}
&\multirow{2}{*}{OK-VQA}
\\ 

D1
&D2
&$\textcolor{Green}{\Smat}$
&$\Amat^{\text{dn}}$
&$\Amat^{\text{up}}$
&(M)
&
&
&
\\

\hline

\rowcolor{gray!20}
\multicolumn{5}{c|}{Finetuning}
&224.54
&46.87$~\std{0.57}$
&34.22$~\std{0.59}$
&16.65$~\std{1.02}$
\\

\hline 

$\checkmark$
&
&
&
&
&2.45
&48.15$~\std{0.89}$
&36.42$~\std{0.64}$
&17.34$~\std{1.59}$
\\ 

$\checkmark$
&
&$\checkmark$
&$\checkmark$
&
&1.55
&47.79$~\std{1.11}$
&36.59$~\std{0.64}$
&\textbf{18.77}$~\std{0.99}$
\\

\rowcolor{DeepSkyBlue!20}
\multicolumn{1}{l}{$\checkmark$}
&$\checkmark$
&$\checkmark$
&$\checkmark$
&
&0.87
&\textbf{48.26}$~\std{0.56}$ 
&\textbf{36.75}$~\std{0.55}$
&18.58$~\std{1.42}$
\\ 

\hline

$\checkmark$
&$\checkmark$
&
&
&
&1.37
&47.67$~\std{1.13}$
&36.22$~\std{0.89}$
&17.65$~\std{2.38}$
\\ 

$\checkmark$
&$\checkmark$
&$\checkmark$
&
&
&1.36
&48.05$~\std{0.99}$
&\textbf{36.76}$~\std{0.78}$
&17.02$~\std{1.70}$
\\ 

$\checkmark$
&$\checkmark$
&
&
&$\checkmark$
&0.83
&47.30$~\std{0.92}$
&36.05$~\std{1.05}$
&17.27$~\std{0.63}$
\\ 

$\checkmark$
&$\checkmark$
&$\checkmark$
&
&$\checkmark$
&0.82
&47.83$~\std{0.65}$
&36.39$~\std{0.84}$
&17.75$~\std{1.36}$
\\ 

$\checkmark$
&$\checkmark$
&
&$\checkmark$
&
&0.88
&47.78$~\std{1.20}$
&36.57$~\std{0.81}$
&18.07$~\std{1.73}$
\\ 

\rowcolor{DeepSkyBlue!20}
\multicolumn{1}{c}{$\checkmark$}
&$\checkmark$
&$\checkmark$
&$\checkmark$
&
&0.87
&\textbf{48.26}$~\std{0.56}$  
&36.75$~\std{0.55}$
&\textbf{18.58}$~\std{1.42}$
\\ 

\hline
\end{tabularx}
}
\vspace{-3mm}
\caption{\textbf{Ablation on weight decomposition (WD) and weight sharing (WS).}
D1: the decomposition of expert weights in MixPHM with PHM. 
D2: the further low-rank reparameterization of the decomposed weights.}
\label{tab:abl_param_redundancy}
\vspace{-3mm}
\end{table}

%% file: table/t5_dis_generalization.tex
\begin{table}[!t]
\footnotesize
\centering 
\setlength{\tabcolsep}{1.mm}{
\begin{tabularx}{\linewidth}{@{}llccccc@{}}
\toprule

\multirow{2}{*}{VLMs} 
&\multirow{2}{*}{Method} 
&\#Param 
&\multicolumn{4}{c}{\#Sample}
\\
& &(M) 
&$N_{\mathcal{D}}$=16 
&$N_{\mathcal{D}}$=64 
&$N_{\mathcal{D}}$=500 
&$N_{\mathcal{D}}$=1000
\\
\midrule

X-VLM
&Finetuning
&294 &26.63 &30.45 &38.96 &43.92 
\\
\cite{zeng2022multi}
&MixPHM
&0.66 &\textbf{27.54} &\textbf{31.80} &\textbf{41.05} &\textbf{48.06} 
\\

\midrule

BLIP
&Finetuning
&385 &27.01 &30.05 &37.00 &42.22
\\
\cite{li2022blip}
&MixPHM
&0.87 &\textbf{29.17} &\textbf{32.09} &\textbf{41.80} &\textbf{46.78} 
\\

\midrule

OFA$_{\text{Base}}$
&Finetuning
&180 &27.48 &31.75 &42.99 &46.81
\\
\cite{wang2022ofa}
&MixPHM
&0.70 &\textbf{28.46} &\textbf{33.00} &\textbf{45.88} &\textbf{50.01} 
\\
\bottomrule
\end{tabularx}
}
\vspace{-3mm}
\caption{\textbf{Experimental results of MixPHM on other pretrained VLMs.}
We report the average VQA-Score across five seeds on VQA v2 validation set under different low-resource settings. 
}
\label{tab:generalization}
\vspace{-3mm}
\end{table}

%% file: section/supp.tex
\noindent \textbf{\Large Appendix} \\ 
\appendix
\label{sec:appendix}

In the Appendix, we provide some supplementary material for our experiments. 
Specifically, \cref{sec:imp_abl} explores the impact of different routing mechanisms and hyperparameters on MixPHM. 
\cref{sec:imp_vis} presents some visualization results of our method.  
\cref{sec:imp_xvlm} provides additional comparisons between parameter-efficient tuning methods using pretrained X-VLM on VQA v2. 
\cref{sec:imp_detail} describes more implementation details.

\section{Ablation Study and Parameter Analysis} 
\label{sec:imp_abl}

In this section, using pretrained VL-T5~\cite{cho2021unifying} as the underlying pretrained VLMs, we conduct additional ablation experiments on routing mechanisms and hyperparameter analysis on VQA v2, GQA, and OK-VQA with $N_{\mathcal{D}}=64$.

\pgraph{Impact of Routing Mechanisms} 
In MixPHM, in addition to performance, routing mechanisms also affect the training speed, \ie, T/Itr (s). 
To analyze the impact of different routing strategies on performance and speed, we first introduce two random routing methods, \ie, token-level and sentence-level routing~\cite{fedus2021switch}. 
In addition, we develop a simple representation-based rounding by averaging the outputs of all PHM-experts in each MixPHM. 
\Cref{tab:abl_routing} shows that random routing mechanism is the fastest and has the best performance on both VQA v2 and OK-VQA.

\pgraph{Impact of Hyperparameters} 
To investigate the impact of different hyperparameters on MixPHM, we conduct experiments by varying $N_e$, $d_r$, $d_k$, and $n$. 
More specifically, we consider the following settings: $N_e\in \{1,2,4,8,12\}$, $d_r \in \{48, 64, 96, 192\}$, $d_k \in \{1, 8, 16, 24\}$, and $n \in \{2, 4, 8, 16\}$. 
The results in \Cref{tab:abl_supp} show that changing these hyperparameters has only a slight impact on the performance of MixPHM. 
In addition, the performance of MixPHM with different hyperparameters always outperforms full finetuning. 
This suggests that the performance improvement brought by MixPHM does not significantly depend on the hyperparameter selection.

\begin{figure*}[!t]
\begin{subfigure}[b]{0.328\linewidth}
\centering
\includegraphics[width=.935\linewidth]{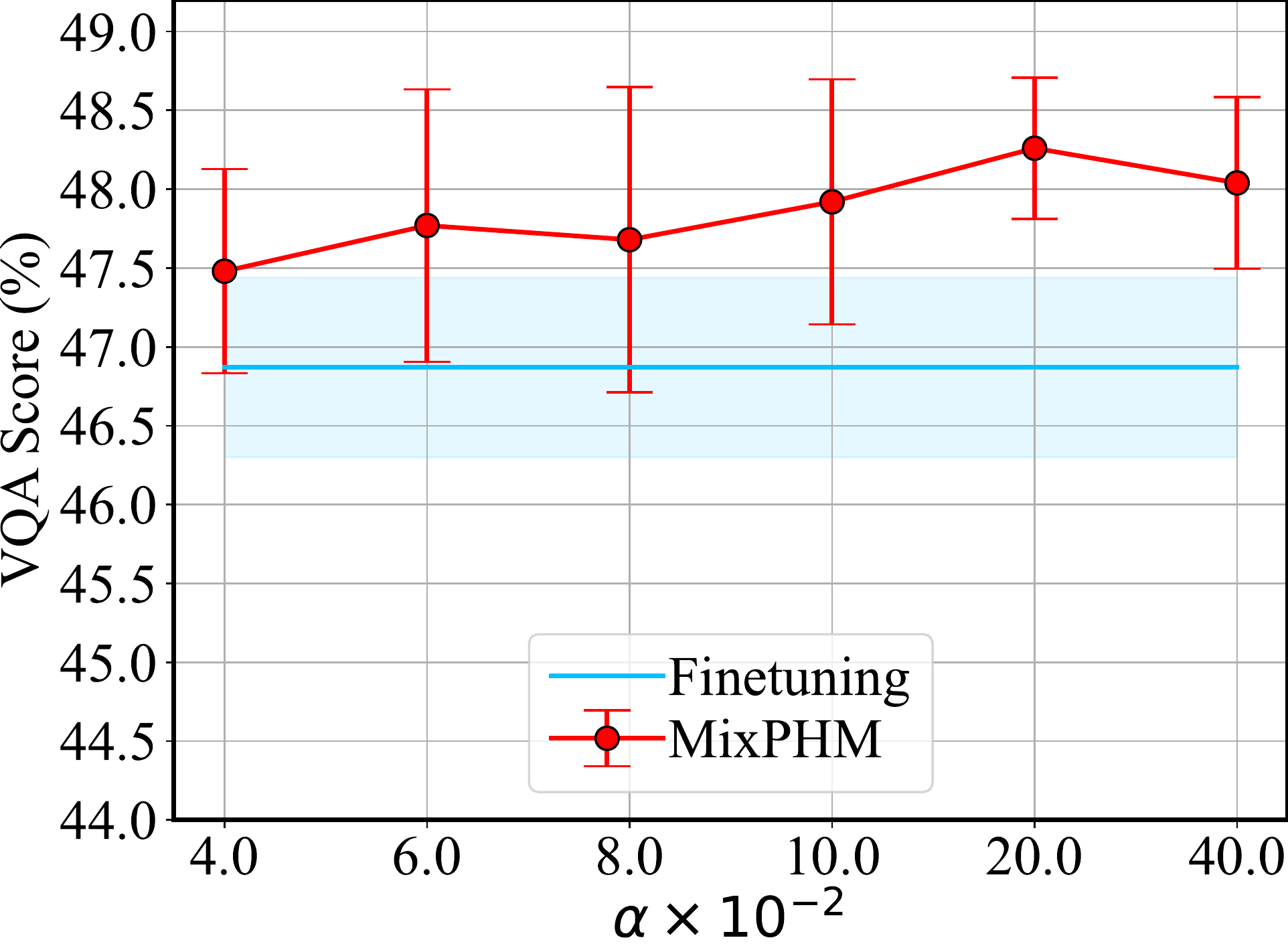}
\subcaption{VQA v2}
\label{fig:abl_hp_alpha_vqav2}
\end{subfigure}
\begin{subfigure}[b]{0.328\linewidth}
\centering
\includegraphics[width=.935\linewidth]{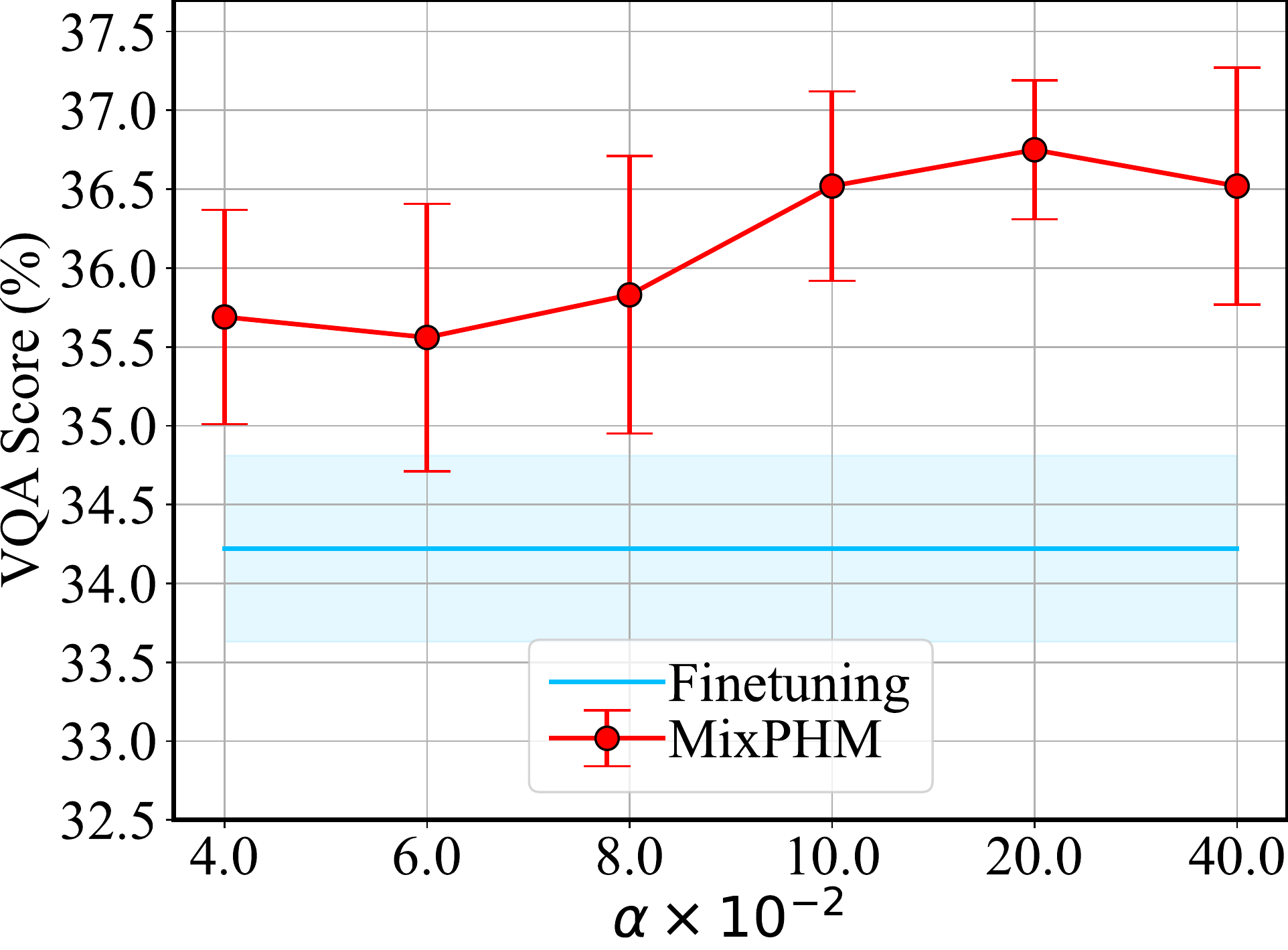}
\subcaption{GQA}
\label{fig:abl_hp_alpha_gqa}
\end{subfigure}
\begin{subfigure}[b]{0.328\linewidth}
\centering
\includegraphics[width=.935\linewidth]{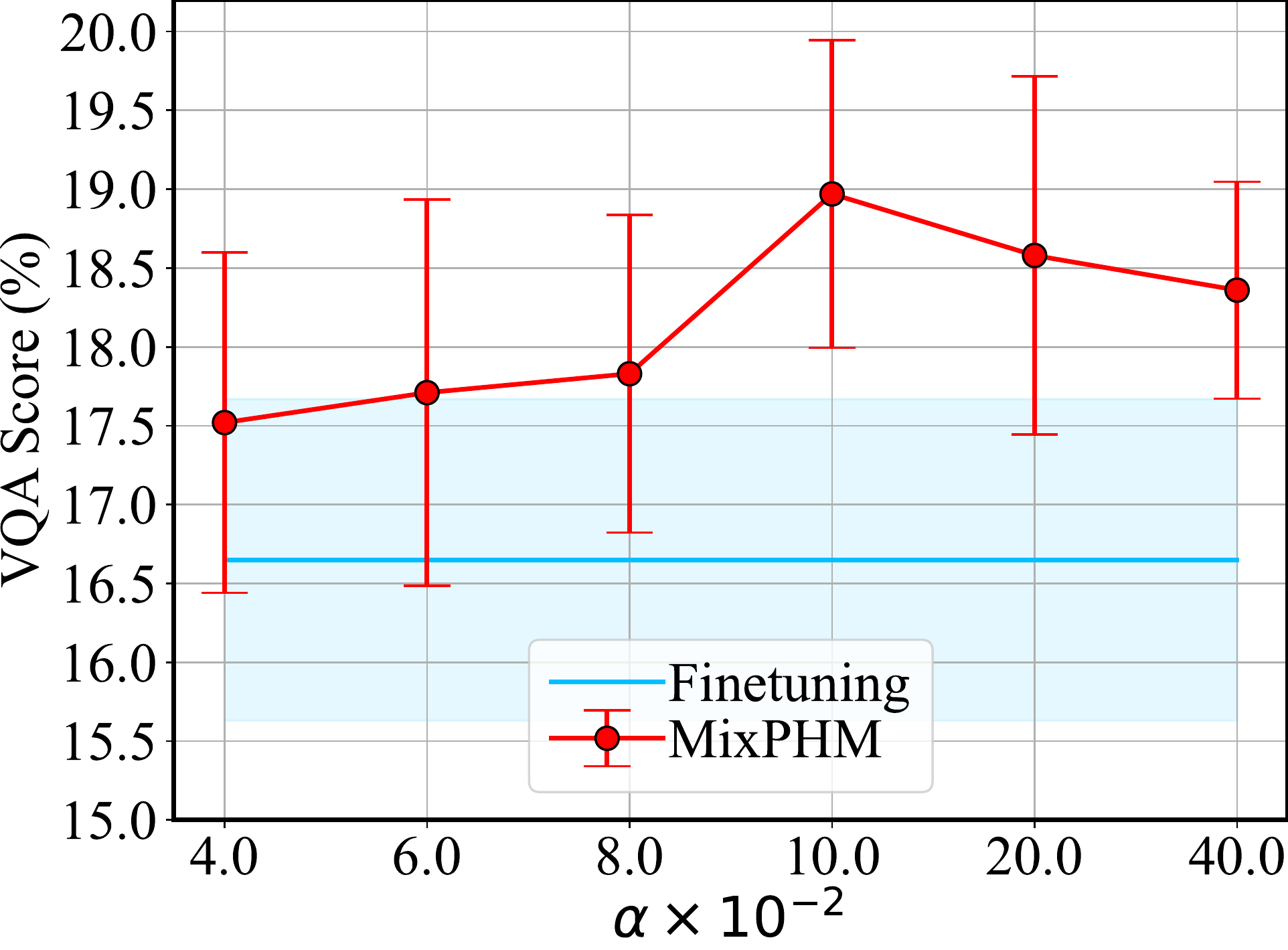}
\subcaption{OK-VQA}
\label{fig:abl_hp_alpha_okvqa}
\end{subfigure}
\vspace{-2mm}
\caption{\textbf{The average VQA-Score with standard deviation across five seeds as $\alpha$ varies.}
}
\label{fig:abl_hp_alpha}
\end{figure*}

\pgraph{Impact of $\alpha$} 
When tuning pretrained VLMs with MixPHM, $\alpha$ controls the trade-off between redundancy regularization and generative modeling loss. 
To investigate the impact of $\alpha$ on MixPHM, we perform experiments with different values of $\alpha$, \ie, $\alpha \in \{0.04, 0.06, 0.08, 0.1, 0.2, 0.4\}$. 
\Cref{fig:abl_hp_alpha} illustrates the curve of VQA-Score as $\alpha$ increases. 
We observe that varying $\alpha$ within a certain range $[0.04, 0.4]$ does not hinder the advantage of MixPHM over full finetuning. 
In addition, according to the results on three datasets, we empirically set $\alpha$ to 0.2.

\input{table/t6_abl_routing.tex}
\input{table/t7_abl_hyperparam.tex}

\begin{figure*}[!t]
\centering
\includegraphics[width=1.0\linewidth]{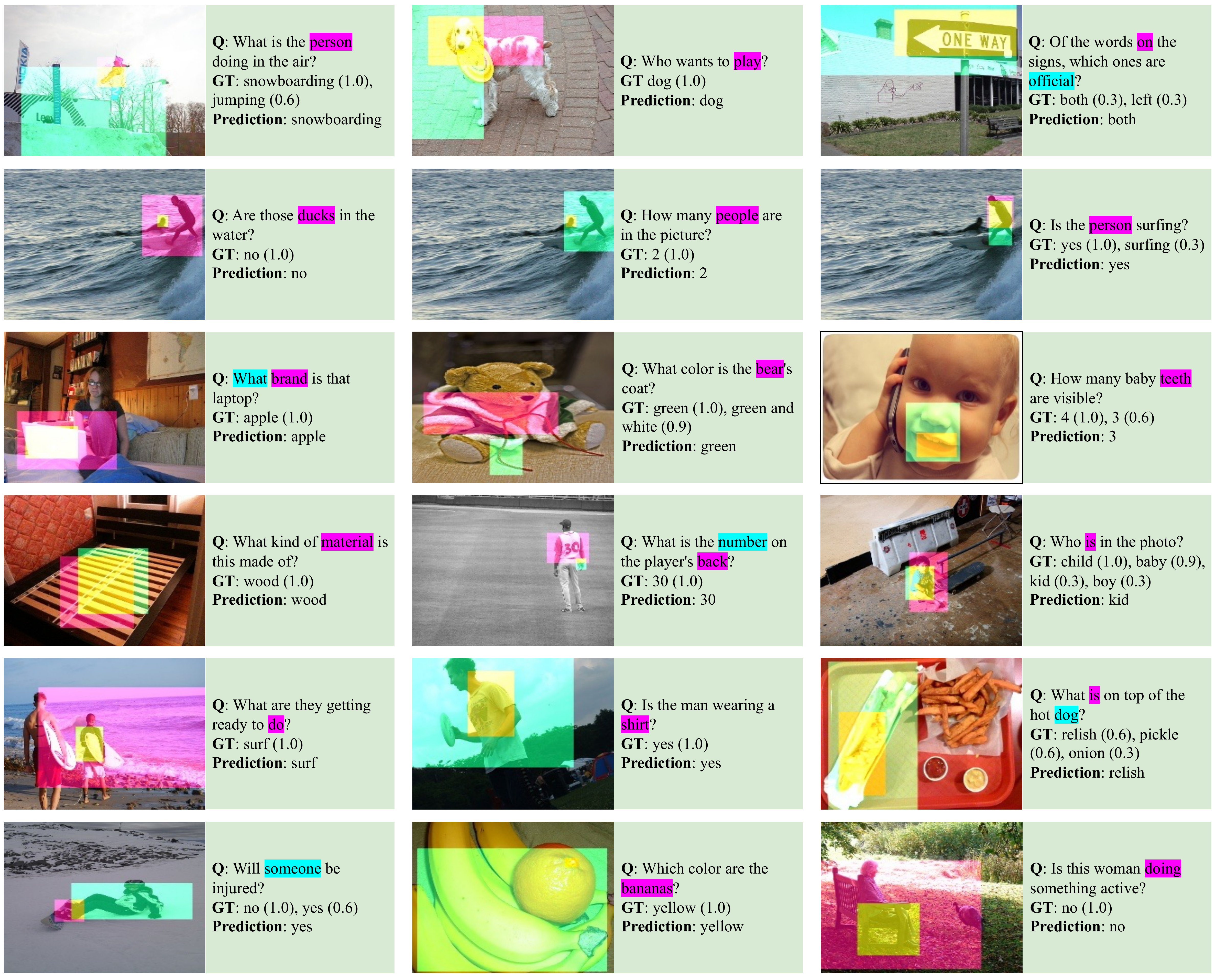}
\vspace{-7mm}
\caption{\textbf{Qualitative results on VQA v2 validation set.} 
The \textbf{Prediction} is generated by the VL-T5 tuned with the proposed MixPHM. 
\textbf{GT} denotes the annotated answer and the corresponding score. 
We visualize the top-down attention~\cite{anderson2018bottom} of images and mark the task-relevant tokens of questions for the \textcolor{magenta}{first} and \textcolor{Cyan2}{second} highest attention scores. 
}
\label{fig:vis_attn}
\end{figure*}

\input{table/t9_pet_xvlm.tex}

\section{Visualization Results}  
\label{sec:imp_vis}

We visualize some examples of the proposed MixPHM. 
As depicted in \Cref{fig:vis_attn}, these predictions are generated by the VL-T5 tuned with MixPHM on VQA v2 with $N_{\mathcal{D}}=64$. 
In addition, we visualize the top-down attention~\cite{anderson2018bottom} for images and mark the top two task-relevant tokens for questions. 
More specifically, we follow a recent work~\cite{jiang2022finetuning} to compute an attention score between task-relevant representations and visual input features obtained using bottom-up Faster R-CNN~\cite{anderson2018bottom} and visualize the top-down attention for the first and second highest scores. 
Analogously, we compute the score between task-relevant representations and linguistic embeddings of questions and mark the tokens for the first and second highest scores. 
Figure~\ref{fig:vis_attn} qualitatively shows that our MixPHM can generate consistent and question-relevant visual and textual attentions in the process of answer prediction.

\section{Results with Pretrained X-VLM}
\label{sec:imp_xvlm} 

As a supplement to the results in Table 5 of the main paper, we utilize pretrained X-VLM~\cite{zeng2022multi} as a representative and compare our methods with state-of-the-art parameter-efficient tuning methods on VQA v2 validation set. 
The key hyperparameter settings for these parameter-efficient methods are the same as those in \Cref{tab:supp_configure}. 
The conclusions that we observe in \Cref{tab:supp_vqa_pet_xvlm} are consistent with Table 5, \ie, our method consistently outperforms existing parameter-efficient tuning methods when using other pretrained VLMs, which further demonstrates the generalization capability of MixPHM.

\section{Implementation Details} 
\label{sec:imp_detail}

For parameter-efficient tuning methods, we search the bottleneck dimension $d_r$ from $\{48, 64, 96, 192\}$ for all adapter-based methods (\ie, MixPHM, AdaMix, Pfeiffer, Houlsby and Compacter), the number of experts $N_e$ from $\{1, 2, 4, 8, 12\}$ for MixPHM and AdaMix, the rank dimension $d_r$ (for MixPHM and Compacter), $r$ (for LoRA) from $\{1, 8, 16, 24\}$, as well as the number of summations of Kronecker product $n$ from $\{2, 4, 8, 16\}$ for MixPHM and Compacter. 
\Cref{tab:supp_configure} presents the final configuration of the hyperparameters used in our experiments. 
For MixPHM, we set the trade-off factor $\alpha$ to 0.2. 

All methods are implemented using Pytorch~\cite{paszke2019pytorch} on an NVIDIA GeForce RTX 3090Ti GPU. 
In addition, we also perform a grid search to select the best learning rate from $\{5\times10^{-5}, 1\times10^{-4}, 5\times10^{-4}, 1\times10^{-3}, 5\times10^{-3}\}$. 
The batch size and the number of epochs are set to 16 and 1000, respectively. 
We utilize AdamW optimizer~\cite{loshchilov2017decoupled} and the early stopping strategy with a patience of 200 non-increasing epochs, where the stopping metric is the VQA-Score on the development set $\mathcal{D}_{\text{dev}}$ of datasets.

\input{table/t8_config.tex}


%% file: table/t6_abl_routing.tex
\begin{table}[t]
\footnotesize 
\centering 
\setlength{\tabcolsep}{1.4mm}{
\begin{tabularx}{\linewidth}{lcccc}
\toprule
\multirow{1}{*}{Method}
&T/Itr (s)
&\multicolumn{1}{c}{VQA v2}
&\multicolumn{1}{c}{GQA}
&\multicolumn{1}{c}{OK-VQA}
\\ 
\midrule

MixPHM-Token
&0.693
&47.67$~\std{0.92}$ 
&36.23$~\std{0.89}$ 
&17.77$~\std{0.89}$
\\

MixPHM-Sent
&0.683
&47.69$~\std{0.99}$ 
&36.13$~\std{0.86}$ 
&17.83$~\std{1.32}$
\\

MixPHM-Rep
&0.675
&48.00$~\std{0.95}$ 
&\textbf{36.77$~\std{0.55}$} 
&18.25$~\std{1.46}$
\\ 

\rowcolor{DeepSkyBlue!20}
MixPHM
&0.668
&\textbf{48.26$~\std{0.56}$}
&\text{36.75$~\std{0.55}$}
&\textbf{18.58$~\std{1.42}$}
\\ 
\bottomrule
\end{tabularx}
}
\vspace{-3mm}
\caption{\textbf{Ablation on different routing mechanisms with $N_{\mathcal{D}}=64$.} 
T/Itr (s) is the average tuning time for each iteration. 
}
\label{tab:abl_routing}
\end{table}

%% file: table/t7_abl_hyperparam.tex
\begin{table}[!t]
\footnotesize 
\centering 
\setlength{\tabcolsep}{2.08mm}{
\begin{tabularx}{\linewidth}{lccccc}
\toprule
\multicolumn{2}{c}{HP}
&\#Param
&\multicolumn{1}{c}{VQA v2}
&\multicolumn{1}{c}{GQA}
&\multicolumn{1}{c}{OK-VQA}
\\ 
\midrule

\rowcolor{gray!20}
\multicolumn{2}{l}{Finetuning}
&224.54
&46.87$~\std{0.57}$
&34.22$~\std{0.59}$
&16.65$~\std{1.02}$
\\

\midrule

&1
&0.34
&47.30$~\std{0.97}$ 
&36.30$~\std{0.83}$ 
&17.59$~\std{0.97}$
\\

&2
&0.52
&47.90$~\std{0.65}$ 
&\textbf{36.88$~\std{0.75}$} 
&18.08$~\std{1.28}$
\\

\rowcolor{DeepSkyBlue!20}
\cellcolor{white}
&4
&0.87
&\textbf{48.26$~\std{0.56}$}
&\text{36.75$~\std{0.55}$} 
&\textbf{18.58$~\std{1.42}$}
\\ 

&8
&1.59
&48.09$~\std{0.67}$  
&36.50$~\std{0.81}$ 
&\text{18.51$~\std{1.29}$}
\\

\multirow{-5}{*}{{$N_e$}}
&12
&2.30
&47.80$~\std{0.72}$  
&36.30$~\std{0.80}$ 
&18.43$~\std{1.50}$
\\ 

\midrule

&48
&0.86
&\textbf{48.36}$~\std{0.97}$ 
&36.36$~\std{0.32}$ 
&18.05$~\std{0.85}$
\\

\rowcolor{DeepSkyBlue!20}
\cellcolor{white}
&64
&0.87
&{48.26$~\std{0.56}$}
&\textbf{36.75$~\std{0.55}$}
&\textbf{18.58$~\std{1.42}$}
\\ 

&96
&0.91
&48.05$~\std{0.82}$
&36.36$~\std{0.51}$
&18.39$~\std{0.96}$
\\ 

\multirow{-4}{*}{{$d_r$}}
&192
&1.00
&47.97$~\std{1.17}$
&36.37$~\std{0.82}$
&18.26$~\std{1.20}$
\\

\midrule

&1
&0.18
&47.87$~\std{0.73}$ 
&35.74$~\std{0.70}$ 
&17.04$~\std{0.81}$
\\

\rowcolor{DeepSkyBlue!20}
\cellcolor{white}
&8
&0.87
&{48.26$~\std{0.56}$}
&\textbf{36.75$~\std{0.55}$}
&{18.58$~\std{1.70}$}
\\ 

&16
&1.67
&\textbf{48.35}$~\std{1.14}$ 
&36.62$~\std{0.35}$ 
&18.22$~\std{1.36}$
\\

\multirow{-4}{*}{{$d_k$}}
&24
&2.47
&48.07$~\std{1.12}$ 
&36.42$~\std{0.52}$
&\textbf{18.79}$~\std{1.18}$
\\ 

\midrule
&2
&0.87
&48.17$~\std{0.93}$ 
&36.53$~\std{0.32}$ 
&18.43$~\std{0.75}$
\\

\rowcolor{DeepSkyBlue!20}
\cellcolor{white}
&4
&0.87
&\textbf{48.26$~\std{0.56}$}
&\textbf{36.75$~\std{0.55}$}
&\textbf{18.58$~\std{1.42}$}
\\ 

&8
&0.87
&47.97$~\std{1.08}$ 
&36.37$~\std{0.56}$ 
&17.41$~\std{1.05}$
\\

\multirow{-4}{*}{{$n$}} 
&16
&0.88
&46.65$~\std{1.10}$ 
&35.46$~\std{0.55}$
&17.52$~\std{0.63}$
\\ 

\bottomrule
\end{tabularx}
}
\vspace{-3mm}
\caption{\textbf{Impact of hyperparameters (HP) on MixPHM.} 
$N_e$: the number of PHM-experts, $d_r$: bottleneck dimension, $d_k$: rank dimension, $n$: the number of summations of Kronecker product. 
}
\label{tab:abl_supp}
\end{table}

%% file: table/t9_pet_xvlm.tex
\begin{table*}[!t]
\centering
\small
\setlength{\tabcolsep}{2.73mm}{
\begin{tabularx}{\linewidth}{lcccccccc}
\toprule
\multirow{2}{*}{Method} 
&\multicolumn{2}{c}{\#Param} 
&\multicolumn{6}{c}{\#Sample}
\\


&\multicolumn{1}{c}{(M)} 
&\multicolumn{1}{c}{(\%)} 

&$N_{\mathcal{D}}$=16 
&$N_{\mathcal{D}}$=32 
&$N_{\mathcal{D}}$=64 
&$N_{\mathcal{D}}$=100 
&$N_{\mathcal{D}}$=500 
&$N_{\mathcal{D}}$=1,000 
\\ 

\midrule


\rowcolor{gray!30}
Finetuning &293.48 
&100\%
&26.63$~\std{0.98}$
&29.33$~\std{1.68}$
&30.45$~\std{1.80}$
&31.48$~\std{1.57}$
&38.96$~\std{1.56}$  
&43.92$~\std{1.22}$
\\ 

BitFit~\cite{zaken2022bitfit} 
&0.29  
&0.13\%
&25.48$~\std{3.81}$
&28.90$~\std{1.14}$
&30.73$~\std{1.18}$
&31.92$~\std{1.14}$
&36.77$~\std{1.32}$
&40.77$~\std{0.79}$
\\

LoRA~\cite{hu2022lora}
&0.37 
&0.13\%
&25.31$~\std{1.50}$
&26.91$~\std{3.09}$
&30.52$~\std{1.67}$ 
&\text{\color{blue}{31.97$~\std{1.11}$}}
&36.13$~\std{1.12}$ 
&40.49$~\std{0.87}$
\\ 

Compacter~\cite{karimi2021compacter} 
&0.25
&0.09\%
&25.69$~\std{2.34}$
&28.04$~\std{1.63}$
&28.10$~\std{2.06}$ 
&31.35$~\std{0.34}$
&35.91$~\std{0.65}$
&40.44$~\std{0.77}$
\\

Houlsby~\cite{houlsby2019parameter}
&3.57 
&1.20\%
&26.54$~\std{2.57}$  
&\text{\color{blue}{29.34$~\std{2.25}$}}
&\text{\color{blue}{30.74$~\std{1.20}$}}
&31.71$~\std{1.43}$
&\text{\color{blue}{38.48$~\std{0.91}$}}
&41.96$~\std{0.72}$
\\

Pfeiffer~\cite{pfeiffer2020adapterfusion}
&1.78
&0.60\%
&\text{\color{blue}{26.57$~\std{2.00}$}}
&28.46$~\std{1.74}$
&29.22$~\std{2.56}$
&31.95$~\std{1.34}$
&37.39$~\std{0.73}$
&40.96$~\std{1.09}$
\\

AdaMix~\cite{wang2022adamix} 
&4.44 
&1.49\%
&26.11$~\std{1.58}$  
&28.91$~\std{1.36}$
&30.71$~\std{2.05}$
&31.15$~\std{1.26}$
&\text{\color{blue}{38.48$~\std{1.53}$}}
&\text{\color{blue}{43.26$~\std{0.85}$}}
\\ 

\textbf{MixPHM} 
&0.66
&0.22\%
&\text{\color{red}{27.54$~\std{1.52}$}}
&\text{\color{red}{30.65$~\std{1.09}$}}
&\text{\color{red}{31.80$~\std{1.61}$}}
&\text{\color{red}{32.58$~\std{1.09}$}}
&\text{\color{red}{41.05$~\std{1.22}$}}
&\text{\color{red}{48.06$~\std{0.64}$}}
\\ 

\bottomrule
\end{tabularx}
}
\vspace{-3mm}
\caption{
\textbf{Experimental results with pretrained X-VLM.} 
The average VQA-Score with standard deviation across 5 different seeds are evaluated on VQA v2 validation set. 
The \text{\color{red}{best}} and \text{\color{blue}{second best}} parameter-efficient tuning methods are highlighted. 
The number of tuned parameters and the percentage of tuned parameters relative to X-VLM (\ie, 293.48M) are reported. 
}
\label{tab:supp_vqa_pet_xvlm}
\end{table*}

%% file: table/t8_config.tex
\begin{table}[!t]
\vspace{2mm}
\footnotesize 
\centering 
\setlength{\tabcolsep}{3.mm}{
\begin{tabularx}{\linewidth}{lcc}
\toprule
\multirow{1}{*}{Method} 
&\multicolumn{1}{c}{Learning rate}
&\multicolumn{1}{c}{Configuration}
\\ 
\midrule

Finetuning
&$5 \times 10^{-5}$ 
&---
\\ 

BitFit
&$5 \times 10^{-5}$ 
&---
\\

LoRA
&$5 \times 10^{-5}$ 
&$r=4$
\\

Compacter
&$5 \times 10^{-3}$ 
&$d_r=64$, $d_k=8$, $n=4$
\\

Houlsby
&$5 \times 10^{-5}$ 
&$d_r=64$
\\

Pfeiffer
&$5 \times 10^{-5}$ 
&$d_r=64$
\\

AdaMix
&$5 \times 10^{-4}$ 
&$N_e=4$, $d_r=64$ 
\\

MixPHM
&$5 \times 10^{-3}$ 
&$N_e=4$, $d_r=64$, $d_k=8$, $n=4$
\\

\bottomrule
\end{tabularx}
}
\vspace{-3.mm}
\caption{\textbf{Hyperparameter settings for all parameter-efficient tuning methods.} 
$N_e$: the number of experts, $d_r$: bottleneck dimension, $d_k$ and $r$: rank dimension, $n$: the number of summations of Kronecker product. 
}
\label{tab:supp_configure}
\vspace{-3mm}
\end{table}